%% file: main.tex
\documentclass[letterpaper]{article} 
\usepackage[draft]{aaai2026}  
\usepackage{times}  
\usepackage{helvet}  
\usepackage{courier}  
\usepackage[hyphens]{url}  
\usepackage{graphicx} 
\usepackage{natbib}  
\usepackage{caption} 
\usepackage{algorithm}
\usepackage{algorithmic}
\usepackage{newfloat}
\usepackage{listings}
\DeclareCaptionStyle{ruled}{labelfont=normalfont,labelsep=colon,strut=off} 
\floatstyle{ruled}
\newfloat{listing}{tb}{lst}{}
\floatname{listing}{Listing}
\usepackage[most]{tcolorbox}

\usepackage{booktabs}
\usepackage{xcolor}
\usepackage{colortbl}
\usepackage{multirow}
\newcommand*\samethanks[1][\value{footnote}]{\footnotemark[#1]}
\title{Verb Mirage: Unveiling and Assessing Verb Concept Hallucinations in \\ Multimodal Large Language Models}
\author{
Zehao Wang\textsuperscript{\rm 1},~
Xinpeng Liu\textsuperscript{\rm 1,2},~
Yudonglin Zhang\textsuperscript{\rm 1},~
Xiaoqian Wu\textsuperscript{\rm 1},~
Zhou Fang\textsuperscript{\rm 1},~
Yifan Fang\textsuperscript{\rm 1},~\\
Junfu Pu\textsuperscript{\rm 3},~
Cewu Lu$\textsuperscript{\rm 1,2}$\thanks{Corresponding authors.},~
Yong-Lu Li\textsuperscript{\rm 1,2}\samethanks
}
\affiliations{
    \textsuperscript{\rm 1}Shanghai Jiao Tong University\\
    \textsuperscript{\rm 2}Shanghai Innovation Institute\\
    \textsuperscript{\rm 3}ARC Lab, Tencent PCG\\
    \{davidwang200099, rightoverthere, enlighten, joefang, despr0, lucewu, yonglu\_li\}@sjtu.edu.cn,\\
    xinpengliu0907@gmail.com, jevinpu@tencent.com
}

\usepackage{bibentry}
\makeatletter
\DeclareRobustCommand\onedot{\futurelet\@let@token\@onedot}
\def\@onedot{\ifx\@let@token.\else.\null\fi\xspace}
\begin{document}
\input{main_paper}
\clearpage
\include{supp}
\end{document}

%% file: main_paper.tex
\maketitle

\begin{abstract}
Multimodal Large Language Models (MLLMs) have garnered significant attention recently and demonstrate outstanding capabilities in various tasks such as OCR, VQA, captioning, \textit{etc}. However, hallucination remains a persistent issue. While numerous methods have been proposed to mitigate hallucinations, achieving notable improvements, these methods primarily focus on mitigating hallucinations related to \textbf{object/noun concepts}. Verb concepts, which are crucial for understanding human actions, have been largely overlooked. In this paper, to the best of our knowledge, we are the \textbf{first} to investigate the \textbf{verb hallucination} phenomenon of MLLMs from various perspectives. Our findings reveal that most state-of-the-art MLLMs suffer from severe verb hallucination. To assess the effectiveness of existing mitigation methods for object concept hallucination in relation to verb hallucination, we evaluated these methods and found that they do not effectively address verb hallucination. 
To address this issue, we propose a baseline method based on fine-tuning with rich verb knowledge, achieving decent superiority. The experiment results demonstrate that our method significantly reduces hallucinations related to verbs.
\end{abstract}

\begin{links}
     \link{Code}{https://github.com/davidwang200099/Verb_Mirage}
\end{links}

\section{Introduction}

Multimodal Large Language Models (MLLMs)~\cite{zhu2024minigpt,liu2023visual,chen2024internvl,bai2023qwen,li2024monkey} have drawn much attention in both the research and industry communities. Armed with high-quality data, a large number of parameters, and efficient instruction-following fine-tuning, they achieve significant success in various tasks, including OCR, VQA, and image captioning, demonstrating a strong generalization ability.

\begin{figure}[t]
    \centering
    \includegraphics[width=0.9\linewidth]{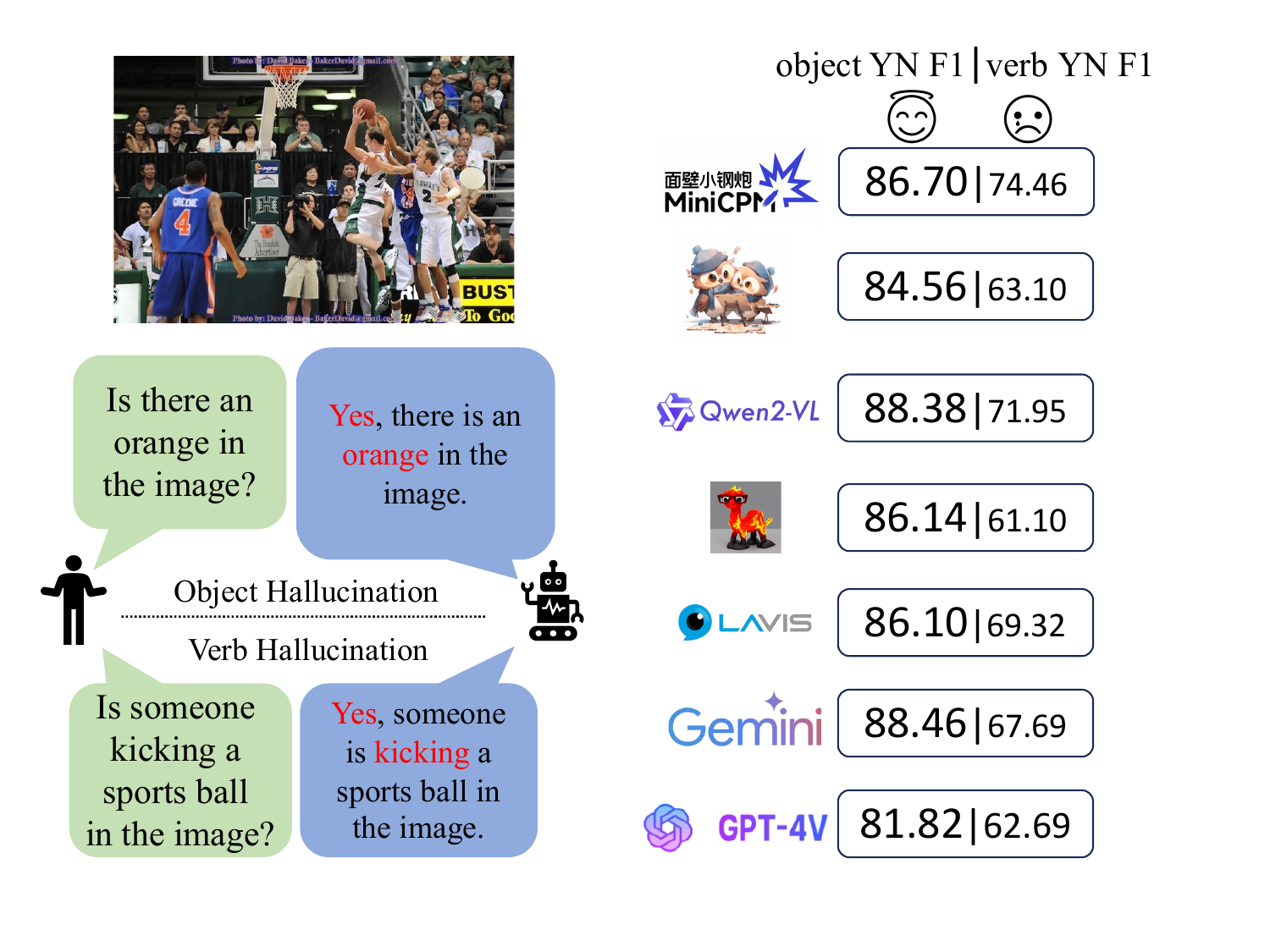} 
    \caption{Besides the well-discussed object hallucination, in this paper, we unveil the severe \textbf{verb hallucination} of state-of-the-art MLLMs with our designed benchmarks. All models show low object hallucination (on POPE) but severe verb hallucination. Gemini-1.5-Flash and GPT-4-Turbo are tested with 100 randomly sampled questions.}
    \label{fig:teaser}
\end{figure}

However, MLLMs' performance improvement could be hindered by hallucination. 
Typically, hallucination~\cite{ji2023survey,liu2024survey} means the output of MLLM contains content against facts, irrelevant or nonsensical given context, such as prompt or multimodal input. 
To test MLLM hallucination in different tasks, many benchmarks~\cite{liu2024mmbench,cai2023benchlmm,fu2023mme} have been made, allowing people to assess MLLMs' abilities in various aspects. 
To mitigate MLLM hallucination, many methods~\cite{huang2024opera,leng2024mitigating,sun2023aligning,yin2023woodpecker,zhou2023analyzing} have been proposed, successfully relieving hallucination to a large extent. 

However, existing benchmarks and methods mainly target hallucination about \textbf{objects/noun-related} concepts. 
\textbf{Action/verb-related} concepts, which are crucial to understanding human events, are overlooked.

To this end, we propose to dig into the verb hallucination problem.
We build the \textbf{first} verb-hallucination-oriented benchmark, which is based on existing datasets~\cite{chao2015hico,sigurdsson2018actor} without the need for extra manual annotations.
As MLLMs are a cooperation of vision and language modalities, we probe MLLM verb hallucination given both different visual inputs and language inputs, covering different query conditions, different imaging conditions, and different semantic conditions. Extensive experiments show that all MLLMs perform poorly on many aspects and thus show severe verb hallucination. 

Moreover, we test existing low-cost hallucination mitigation methods on widely used MLLMs and show that they all fail in mitigating verb hallucination. 
To somewhat relieve verb hallucination, we propose a baseline method based on parameter-efficient fine-tuning with verb structure knowledge. 
Experiments show that our method successfully relieves verb hallucination, but its performance is still far from satisfactory. 
Finally, we explore the reason for verb hallucination and discuss possible future solutions. 

In conclusion, our contributions are: 
\begin{enumerate}
    \item To our knowledge, we point out and analyze MLLM's verb hallucination for the first time and probe this phenomenon under different conditions.
    \item We study the influence of existing training-free and finetuning-based methods on MLLM verb hallucination. We find that fine-tuning is still the most promising way for verb hallucination mitigation.
    \item We probe model behavior from the perspective of vision-language interaction and token uncertainty, and study how well MLLMs learn verbs.
\end{enumerate}

\section{Related Work}

\textbf{MLLM Benchmarks.}
Before the emergence of MLLMs, great efforts have been made to build datasets on tasks like image captioning~\cite{chen2015microsoft,sharma2018conceptual,young2014image}, VQA~\cite{goyal2017making,hudson2019gqa,marino2019ok}, OCR~\cite{singh2019towards}, action recognition and detection~\cite{gu2018ava,tincvpr,djrn,idn,pastanet}, \textit{etc}. 
However, they mainly assess domain-specific models. 
To fully evaluate MLLMs, more benchmarks have been proposed~\cite{liu2024mmbench,ying2024mmtbench,yue2024mmmu,li2024seed,tong2024eyes,li2025the} to test different aspects and subtasks. 
Benchmarks are also proposed to conduct detailed assessments on specific aspects, like BenchLMM~\cite{cai2023benchlmm} for robustness against image styles and MMSpuBench~\cite{ye2024mm} for robustness against spurious correlations.

\textbf{MLLM Hallucination.}
Among the emerging benchmarks, hallucination has become a focus.
Typically, hallucination means that the contents generated by models are untruthful, against facts, or nonsensical~\cite{ji2023survey,liu2024survey,zhang2023siren}. 
Many benchmarks on hallucination have been proposed~\cite{guan2024hallusionbench,kaul2024throne,wang2024haloquest,chen2024multi,zhong2024investigating}. 
Among different types of hallucination, object hallucination~\cite{rohrbach2018object} is deeper explored.
Binary questions are used to probe hallucination about a certain class of objects in POPE~\cite{li2023evaluating}. 
CHAIR score~\cite{rohrbach2018object} is used to measure object hallucination in image captioning. 
In most previous studies, only object concept hallucinations were covered by identifying whether MLLMs refer to objects nonexisting in the image or incorrect attributes of objects. 
Though some other phenomena are studied, such as event hallucination~\cite{jiang2024hal}, relationship hallucination~\cite{wu2024evaluating}, \textit{etc.}, 
verb-related concepts, which are crucial to understanding human actions, were still neglected. 
Event is a complex combination of objects, verbs, adjectives, adverbs, \textit{etc.}, and Hal-eval \cite{jiang2024hal} tests event hallucination by introducing nonexisting objects and can be bypassed by mitigating object hallucination.
Verbs are a kind of relation, but visual relationship also covers \textit{spatial relations, ownership, subject-place relations, attributes}, \textit{etc.}, which account for the majority. In this paper, we focus on the understanding of \textbf{humans existing in the image}, \textit{i.e.}, \textbf{a human-centric problem}. 

\textbf{Hallucination Mitigation.}
Researchers have revealed the reasons for hallucination from many different aspects and proposed hallucination mitigation methods. 
Over-attention to summarizing tokens~\cite{huang2024opera} and language prior~\cite{leng2024mitigating} are recognized to be correlated with object hallucination. 
To mitigate the bias or errors in training data, researchers proposed mitigating bias~\cite{liu2024mitigating,hu2023ciem,gunjal2024detecting} in the dataset or enriching the annotation~\cite{zhai2023halle}. 
Moreover, some suggest post-processing at inference time by adjusting decoding strategy~\cite{leng2024mitigating,chen2424halc} or correcting the output of MLLMs with the help of expert models~\cite{zhou2023analyzing,yin2023woodpecker}. 

\section{Probing on MLLM IO Conditions}
\begin{figure*}[t]
    \centering
    \includegraphics[width=0.8\linewidth]{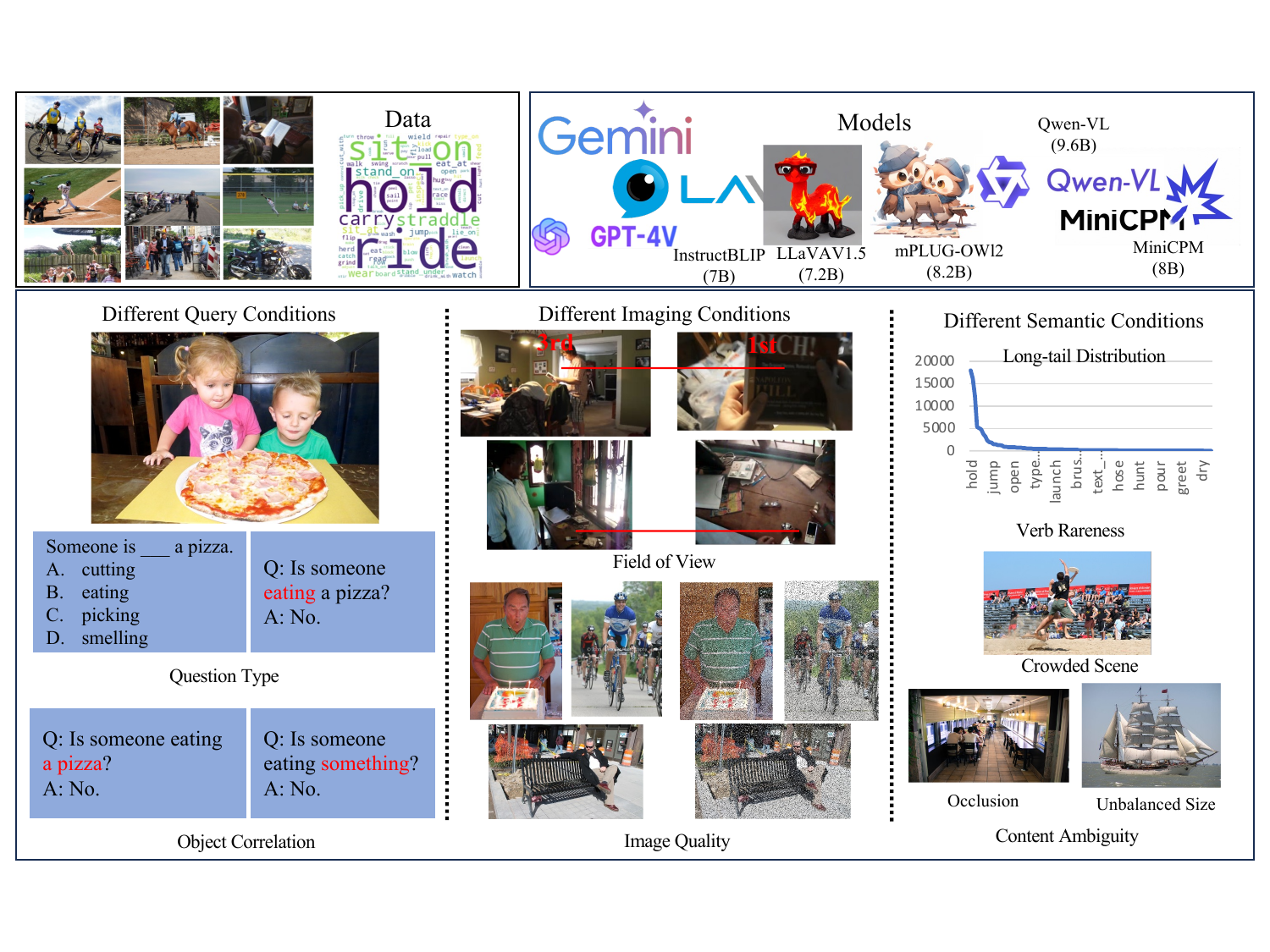} 
    \caption{We probe MLLM verb hallucination from various perspectives, \textit{eg}., question formats, the existence of object correlation, different fields of view, image qualities, verb semantics, and image semantics.}
    \label{fig:method}
\end{figure*}

As shown in Fig.~\ref{fig:method}, we probe verb hallucination from various perspectives: MLLM behavior given different question formats, image qualities, verb semantics, viewpoints, \textit{etc}.

We select HICO~\cite{chao2015hico} and CharadesEgo~\cite{sigurdsson2018actor} as the main datasets for probing verb hallucination. 
HICO contains 47K images with dense annotations. It includes 600 action classes formed by 80 object classes and 117 verb classes. WordNet~\cite{miller1995wordnet} was used to handle synonym and hierarchy problem. It has rich verb labels and is thus suitable for evaluating MLLM verb understanding with minor manual adaptation. 
Meanwhile, CharadesEgo contains 7K videos of daily indoor activities. In each scenario, the same actor is recorded with both an egocentric and an exocentric camera. It contains 157 action classes, each formed by a verb and an object. 

We test several open-sourced and close-sourced MLLMs including InstructBLIP-7B~\cite{dai2023instructblip}, LLaVA-V1.5-7B~\cite{liu2024improved}, mPLUG-Owl2~\cite{ye2024mplug}, Qwen-VL-Chat~\cite{bai2023qwen}, MiniCPM-Llama3-V2.5~\cite{yao2024minicpm}, Qwen2-VL-7B-Instruct~\cite{Qwen2VL}, Molmo-7B-D~\cite{deitke2025molmo}, 
GPT-4~\cite{achiam2023gpt}, Gemini-1.5~\cite{team2023gemini}, \textit{etc}. 
They have different ranks on leaderboards and show outstanding results on benchmarks targeting object concepts. 

\subsection{Probing on Query Conditions}
\subsubsection{Question Formats}
Next, we probe the relation between verb hallucination and QA format.
MLLMs with low hallucinations should give hallucination-free answers, given different question formats. 
Thus, we evaluate verb hallucination using different question formats, including Multiple Choice (MC) questions with only one correct answer each and Yes-or-No (YN) questions. 
Here, we do not introduce free-form image captioning and blank-filling because these two forms require rule-based post-processing on verbs and may lead to severe misclassification errors. 

We aim to verify none other than verb hallucination, so we do our best to omit relevance to object hallucination. 
For each \textit{verb-object} tuple in the labels, we form questions or options by altering the verb and leaving the object unchanged. For YN questions, when building negative samples, we randomly choose verbs that are plausible for the objects but not carried out in the image.
For example, if an image contains a person holding a cup, we may ask MLLM, ``Is someone holding a cup? Is someone washing a cup?'' We regard accuracy, precision, recall, and F1 score as the metrics for hallucination. 
Similarly, when building MC questions, for a sample image, we randomly choose a verb presented in the image and three verbs possibly performable upon objects but not presented. 
We introduce circular evaluation~\cite{liu2024mmbench} and regard accuracy as a metric for MLLM verb hallucination. 
Then, the relevance of object hallucination can be minimized. However, we must point out that as a substantial proportion of verbs are transitive verbs, the influence of objects can not be completely omitted.
The construction of benchmarks is detailed in Suppl. Sec.~{1}.

\subsubsection{Object Correlation}
Sometimes we focus on human interaction with a certain class of object, but sometimes our focus on verbs may be object class agnostic. 
Specifically, we may wonder ``Is someone holding a cup in the image?'' However, sometimes we may also want to know ``Is someone eating something in the image?'' 
Therefore, we test MLLM verb understanding, given reference to objects and not. Among these two conditions, we believe questions without object correlation (\textit{ie}, ``Is someone eating something?'') have less relevance to object hallucination. Still, questions with object correlation are also very practical in daily use.

\subsubsection{Analysis}
\begin{figure*}[t]
\centering
\begin{minipage}[t]{0.6\textwidth}
\centering
{\scriptsize
\begin{tabular}{cccc|ccc|c|c}
\toprule
 & \multicolumn{3}{c}{YN w/ obj} & \multicolumn{3}{c}{YN verb only} &MC w/ obj& MC verb only           \\
Model&acc&prec&recall&acc&prec&recall&acc&acc\\
\midrule
Molmo-7B-D          & 59.16&\cellcolor[HTML]{9de0ff}44.91&\cellcolor[HTML]{f47a75}96.63&46.13&\cellcolor[HTML]{9de0ff}38.39&\cellcolor[HTML]{f47a75}99.19&\textbf{60.64}&56.78\\
Qwen2-VL-7B         & 75.51&58.37&\cellcolor[HTML]{f47a75}93.75 &74.69&57.43&\cellcolor[HTML]{f47a75}95.87&\textbf{71.47}&65.31   \\
MiniCPM-Llama3-V2.5 & 80.91&66.83&85.41 &79.14&63.33&\cellcolor[HTML]{f47a75}90.33&\textbf{66.39}&60.77\\
Qwen-VL-Chat        & 78.06&62.37&87.02  & 79.24&65.09&82.68&\textbf{55.95}&54.57 \\
mPLUG-Owl-2         & 62.94&\cellcolor[HTML]{9de0ff}47.38&\cellcolor[HTML]{f47a75}95.99  & 62.61&\cellcolor[HTML]{9de0ff}47.25&\cellcolor[HTML]{f47a75}94.94&\textbf{63.91}&62.60 \\
LLaVA V1.5          &52.16&\cellcolor[HTML]{9de0ff}40.99&\cellcolor[HTML]{f47a75}97.35&59.16&\cellcolor[HTML]{9de0ff}45.10&\cellcolor[HTML]{f47a75}98.06&\textbf{57.37}&51.00\\
InstructBLIP        & 72.53&55.79&86.77  & 73.82&57.25&87.79 &\textbf{13.48}&6.25\\
\bottomrule
\end{tabular}}
\captionof{table}{Results on YN and MC questions w/ and w/o object reference. {\color[HTML]{f47a75}Red}: high recall. {\color[HTML]{9de0ff}Blue}: low precision. \textbf{Bold}: higher MC acc w/ object reference than w/o object referece.}
\label{object_reference_yon}
\end{minipage}
\hfill
\begin{minipage}[t]{0.35\textwidth}
    \centering
    \vspace{-50px}
    \includegraphics[width=\linewidth,height=0.5\columnwidth]{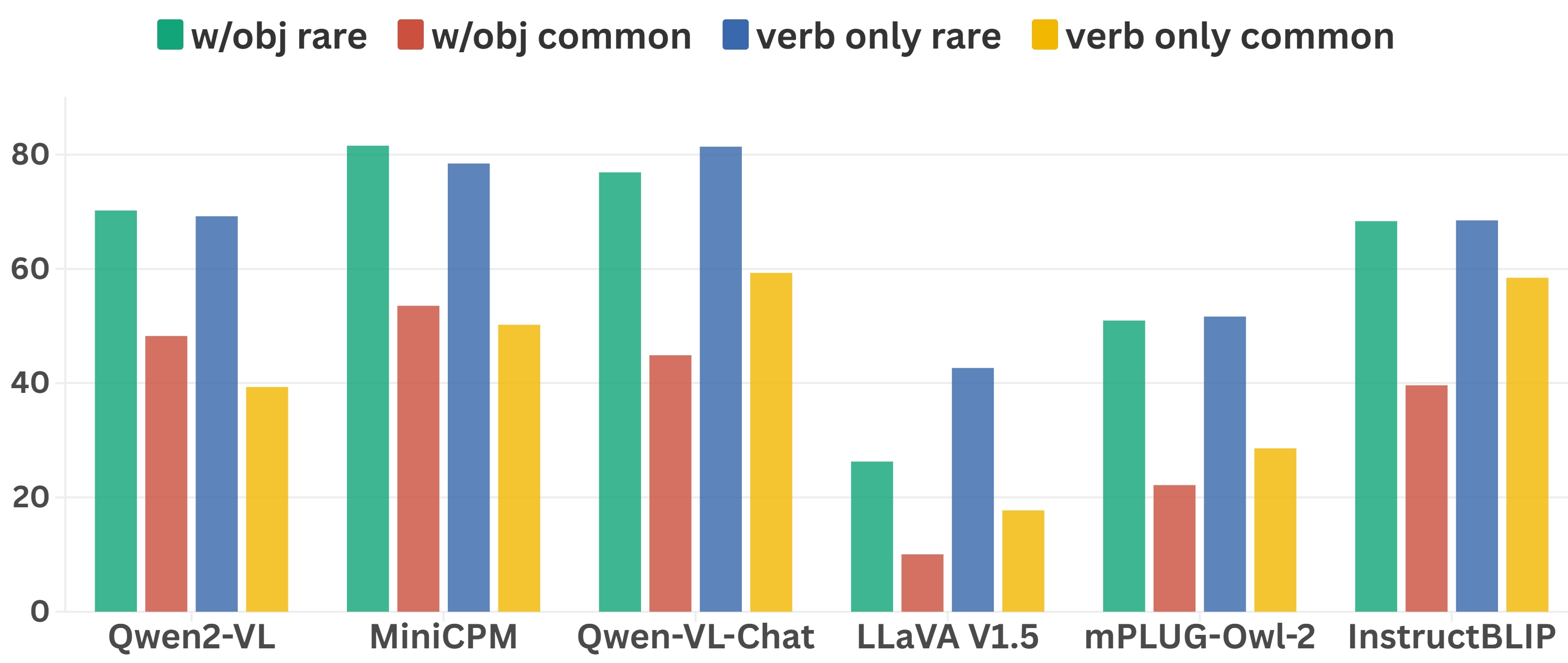}
    \vspace{-15px}
    \captionof{figure}{Comparison of YN questions with correct answer \textit{No} on rare and common subsets.}
    \label{fig:yon_rare_common_subsets}
\end{minipage}
\end{figure*}

The results are shown in Tab.~\ref{object_reference_yon}, giving us rich clues about MLLM verb hallucination.

\begin{figure}
    \centering
    \includegraphics[width=0.9\linewidth,height=0.5\columnwidth]{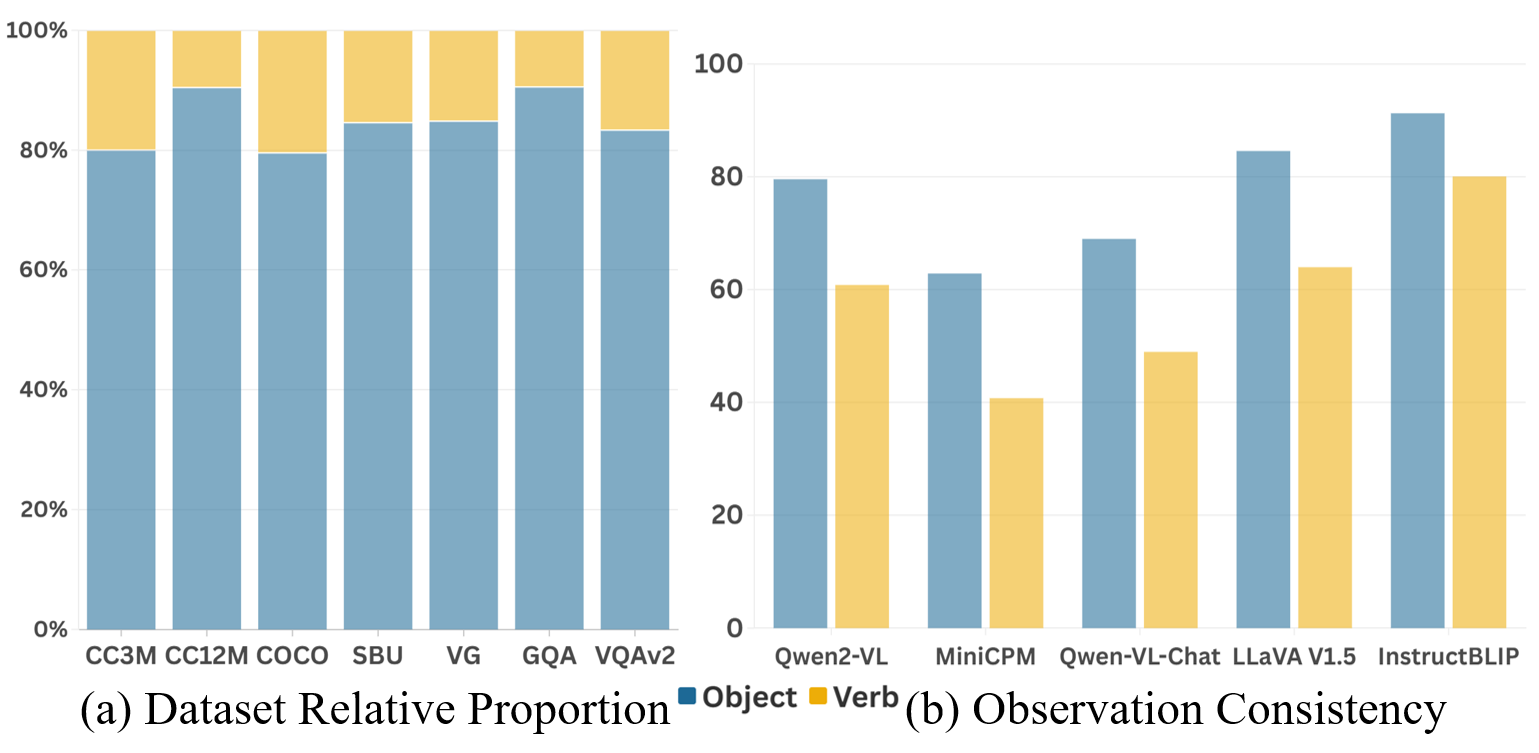}
    \caption{Comparison between objects and verbs.}
    \label{comparison_verb_noun}
\end{figure}

\textbf{Heavy reliance on objects.} 
MLLMs show drastic performance degradation on MC questions without reference to objects. Detailed statistics on YN questions based on object classes referred to in the questions also reveal that MLLM verb understanding relies heavily on object reference. 
We analyze some commonly used datasets~\cite{sharma2018conceptual,changpinyo2021conceptual,lin2014microsoft,ordonez2011im2text,krishna2017visual,hudson2019gqa,goyal2017making} for MLLM pretraining and investigate the number of nouns and verbs in the datasets in Fig.~\ref{comparison_verb_noun}(a). 
We can see that the number of nouns is 4-10 times the number of verbs. One reason for this unbalanced ratio of nouns and verbs is that datasets on action understanding have not attracted enough attention from the MLLM community. Another reason is that there are many more nouns than verbs in English. Research on shortcut learning~\cite{geirhos2020shortcut} also sheds light on the reason for MLLMs' overreliance on nouns.

\textbf{Inability to refuse.} 
All MLLMs have high recall but low precision, meaning that MLLMs tend to give \textit{Yes} whether a certain verb is presented in the image or not. 
Binary questions require MLLMs to have a deep understanding of verb concepts in images, which are more difficult to answer than MC questions, but vitally important in daily use.

\textbf{Similar Bias.}
We ensemble the answers given by three outstanding models and show results in Tab.~\ref{ensemble}. 
There is no substantial improvement over the individual models, showing that the models share similar biases.

\subsection{Probing on Imaging Conditions}
\subsubsection{High-Quality and Low-Quality Images} 
Previous research~\cite{li2022hake,leng2024mitigating} has revealed that visually distorted images can hinder both humans and models from recognizing the contents in images well. However, the relation between visual distortion and verb hallucination is unexplored. 
Do MLLMs hallucinate in the same way when given high-quality images and visually distorted images? Is verb understanding more sensitive to visual distortion than object understanding for MLLMs?
Here, we add pepper-salt noise as a visual distortion to images, each pixel of which is affected at a probability of 75\%.

We evaluate MLLM verb understanding with both high-quality and visually-distorted images and report performance and error consistency following \cite{geirhos2021partial} in Tab.~\ref{pepper_salt_yon}. All tested MLLMs show obvious performance degradation. 
Error consistency in the form of Cohen's Kappa~\cite{cohen1960coefficient} measures MLLM consistency of answers given different visual conditions and provides a guideline for MLLM performance improvement. We can see that some MLLMs with low ranks do not have bad error consistency, but MLLMs with higher ranks show low error consistency. This means that their verb hallucination can be easily induced by visual distortion.

As a control experiment, besides verb understanding, we also built a test set for MLLMs' object understanding with the same set of images. The observation consistency in terms of Cohen's Kappa is reported in Fig.~\ref{comparison_verb_noun}(b). From the result, we can see that MLLM shows much higher inconsistency in verb understanding than object understanding, which means that visual distortion does more harm to verb understanding than object understanding. We attribute this result to the sparsity of verb semantics in pixel space.

In conclusion, \textbf{visual distortion affects both object understanding and verb understanding of MLLMs, but the effect on verb understanding is greater.}
\begin{table*}[t]
\centering
{\scriptsize
\begin{tabular}{ccc|cc|c|c|c|c}
\toprule
        & \multicolumn{5}{c}{YN verb only} &\multicolumn{3}{c}{MC verb only}\\
        & \multicolumn{2}{c|}{w/o Pepper Salt}& \multicolumn{2}{c|}{w/ Pepper Salt}&YN  Err. Cons.&w/o Pepper Salt&w/ Pepper Salt&MC Err. Cons.\\
        &YN acc&YN F1&YN acc&YN F1&&MC acc&MC acc\\
\midrule
MiniCPM-Llama3-V2.5       &  \underline{79.14}&\underline{74.46}& 67.40&57.23&\cellcolor[HTML]{9de0ff}26.12&\underline{60.77}  & 40.50 & \cellcolor[HTML]{9de0ff}37.20\\
Qwen-VL-Chat             &  \underline{79.24}&\underline{72.84}& 66.64&61.88&\cellcolor[HTML]{9de0ff}38.47&\underline{54.57}  & 33.98  &\cellcolor[HTML]{9de0ff}43.38 \\
LLaVA V1.5                &  \underline{59.16}&\underline{61.79}& 51.29&57.35&\cellcolor[HTML]{f47a75}{73.85}&\underline{51.00}  & 49.97   & \cellcolor[HTML]{f47a75}68.37  \\
InstructBLIP              &  \underline{73.82}&\underline{69.30}&71.04&67.02&\cellcolor[HTML]{f47a75}{74.16}&6.25   & \underline{6.34} &\cellcolor[HTML]{f47a75}82.33\\
\bottomrule
\end{tabular}}
\caption{Performance comparison for images w/ and w/o pepper-salt noise. \underline{Underline}: higher performance w/o peppser-salt noise. {\color[HTML]{f47a75}Red}/{\color[HTML]{9de0ff}Blue}: good/bad error consistency (Err. Cons.).}
\label{pepper_salt_yon}
\end{table*}

\begin{table}[t]
\centering
{\scriptsize
\begin{tabular}{cccc|ccc}
\bottomrule
  & \multicolumn{3}{c|}{YN w/ obj}         & \multicolumn{3}{c}{YN verb only}  \\
  &acc&prec&recall&acc&prec&recall\\
\midrule
MiniCPM       &\underline{80.91}&\underline{66.83}&85.41  &79.14&63.33&\underline{90.33}\\
Qwen-VL-Chat  &  78.06&62.37&87.02 &79.24&\underline{65.09}&82.68 \\
InstructBLIP &    72.53&55.79&86.77  & 73.83&57.26&87.80\\
Ensemble & 75.59&58.74&\underline{91.17}  & \underline{79.81}&64.48&89.12\\
\bottomrule
\end{tabular}}
\caption{Ensembled accuracy/precision/recall on YN question w/wo object reference.}
\label{ensemble}
\end{table}

\subsubsection{Egocentric and Exocentric Images}
Recently, MLLMs have shown outstanding capabilities in recognition and reasoning, and a growing body of research has delved into leveraging MLLMs for various tasks in robotics, where egocentric images are widely used. 
Therefore, a study of MLLM's understanding of verbs in egocentric images holds significant importance. To evaluate MLLMs' understanding of verbs in egocentric images and the performance gap between egocentric and exocentric images, we build a small test set using Charades-Ego and conduct experiments on different MLLMs.

\begin{figure}[t]
    \centering
    \includegraphics[width=0.75\linewidth]{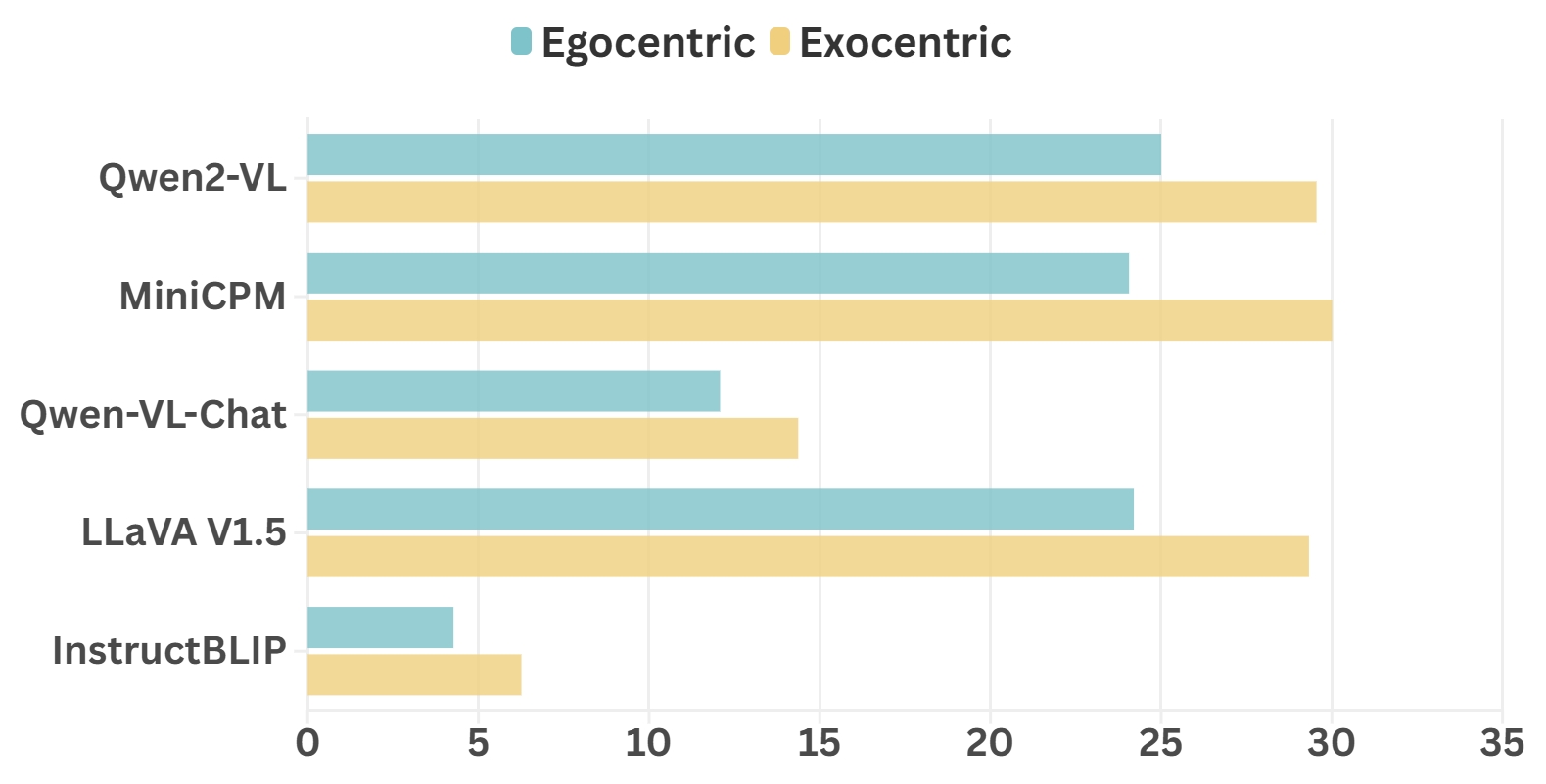}
    \caption{Performance comparison on egocentric and exocentric verb understanding (question type: MCQ).}
    \label{fig:egocentric_mcq}
\end{figure}

Given the same scenario recorded with exocentric and egocentric cameras, we also probe MLLMs' understanding with MC and YN questions. The results are shown in Fig.~\ref{fig:egocentric_mcq} and Tab.~\ref{egocentric_yon}. There is a substantial gap between MLLMs' understanding of exocentric and egocentric images.

In conclusion, \textbf{MLLMs can not understand verb concepts in egocentric images as well as exocentric images. MLLMs are better at exocentric verb understanding.}
\begin{table}[t]
\centering
{\scriptsize
\begin{tabular}{cccc|ccc}

\toprule
View         & \multicolumn{3}{c|}{Egocentric}  & \multicolumn{3}{c}{Exocentric} \\
Model& acc&prec&recall&acc&prec&recall\\
\midrule
Qwen2-VL-7B   &  60.10&60.31&59.12 & \underline{63.01}&\underline{60.98}&\underline{72.25} \\
MiniCPM-Llama3-V2.5 &59.42&62.60&\underline{46.78} &\underline{62.00}&\underline{67.44}&46.39 \\
Qwen-VL-Chat  & 57.06&61.87&36.79  & \underline{60.62}&\underline{64.94}&\underline{46.19}       \\
\bottomrule
\end{tabular}}
\caption{Comparison on egocentric and exocentric verb understanding (question type: YN).}
\label{egocentric_yon}
\end{table}
\subsection{Probing on Semantic Conditions}
\subsubsection{Rare and Common Verbs}
Verbs follow a long-tailed distribution in action datasets because of all sorts of difficulties in the process of dataset collection. However, understanding \textit{rare} and \textit{common} verbs is equally important in real-world applications. 
We hypothesize that MLLMs tend to hallucinate more on rare verbs and try to prove it on existing datasets. 

We divide the negative samples into two subsets: the rare set and the common set. 
In the rare set, the verb in question lies in the \textit{tail} of HICO verb distribution, while in the common set, the verb in question lies in the \textit{head}. Specifically, for YN questions with an object reference, the rare set contains all HOIs whose annotations make up less than 20\% among HOIs relevant to the same object class.
For YN questions without object reference, the common set contains all questions containing \texttt{hold}, \texttt{ride}, \texttt{sit\_on}, \texttt{straddle}, and \texttt{carry}, making up 50\% of the verb annotations in the HICO dataset. 
From Fig.~\ref{fig:yon_rare_common_subsets}, we can see that MLLMs tend to refuse existent rare verbs but accept nonexistent common verbs in images. This phenomenon reveals that the long-tailed distribution of verb annotations limits MLLM verb understanding. How to understand rare verbs remains a problem, and there is a large room for action, data collection, and curation. In conclusion, \textbf{MLLMs can not understand rare verbs as well as common verbs, \textit{i.e.}, long-tail affects a lot}.

\subsubsection{Image Content Ambiguity}
Ambiguity always exists in real-world scenarios. Understanding verbs in crowded or heavily occluded scenarios is important in many fields such as surveillance, social robotics, and visual reasoning. To assess MLLMs' verb understanding given images with ambiguous content, we select images from HICO containing content ambiguities, form an ambiguous subset, and compare them with images with less content ambiguity. 
Specifically, our ambiguous subset contains many contributing factors to ambiguity:
\begin{itemize}
    \item \textbf{Imbalanced human-object relative size.} Imbalanced human-object relative sizes can add difficulties to MLLM's perception of humans and objects. The existence of verbs relies heavily on the accurate perception of humans and objects in images. Potential failures in perception bring great difficulties to the recognition of verbs.
    \item \textbf{Crowded scene.} A highly complicated scene structure can distract MLLMs. To judge the existence of a verb, MLLMs must analyze all humans and objects and draw a comprehensive conclusion. The large number of humans and objects puts heavy burdens on MLLMs.
    \item \textbf{Occlusion.} We can recognize humans and objects according to parts of them and analyze their relationship, even with prominent occlusion. Thus, visual reasoning under occluded scenarios is important in fields like action understanding and robot manipulation. 
\end{itemize}

\begin{figure}
    \centering
    \includegraphics[width=0.75\linewidth]{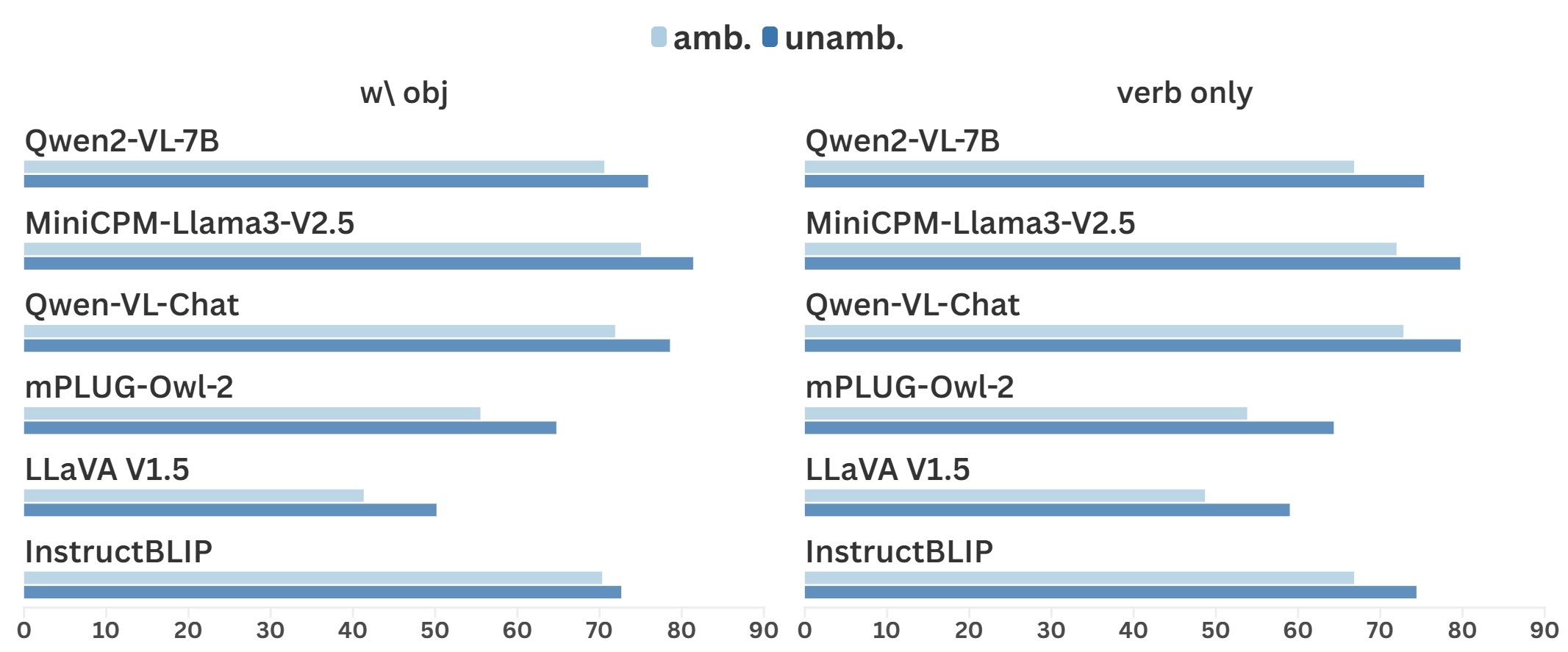}
    \caption{MLLM accuracy on ambiguous (Amb.) and unambiguous (Unamb.) subsets.}
    \label{fig:yon_ambiguous_subsets}
\end{figure}

From Fig.~\ref{fig:yon_ambiguous_subsets}, we have the following finding: \textbf{MLLMs show performance degradation when images contain content ambiguity. How to understand verbs given ambiguous content is an open problem to be solved.}

\section{Probing on Model Behaviors}
In this section, we take LLaVA V1.5 as an example and study the relationship between its behavior and verb hallucination. Our study finds that although verb hallucination shares some commonality with object hallucination, they are fundamentally different.
\subsection{Vision-Language Interaction}
\begin{figure}[t]
    \centering
    \includegraphics[width=0.88\linewidth]{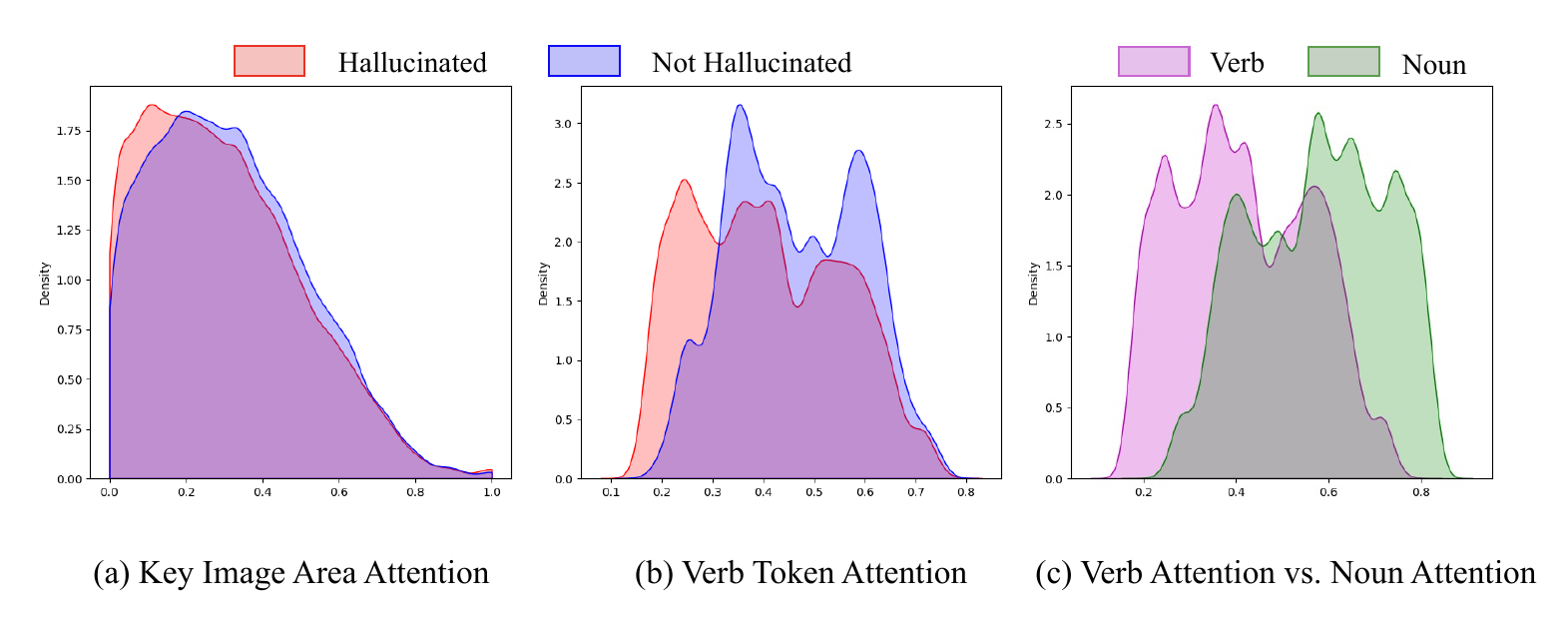}
    \caption{Attention for YN questions with object references.}
    \label{fig:yon_w_object_verbattention}
\end{figure}

\textbf{Key Image Area Attention}.
First, we hypothesize that models pay less attention to key information in images and text when they give hallucinated answers. 
From Fig.~\ref{fig:yon_w_object_verbattention}(a), we can see that there is an obvious distinction of distribution between hallucinated attention and non-hallucinated attention, showing \textbf{a strong correlation between inadequate attention to key areas and hallucination}. 

\begin{figure}[t]
    \centering
    \includegraphics[width=0.9\linewidth]{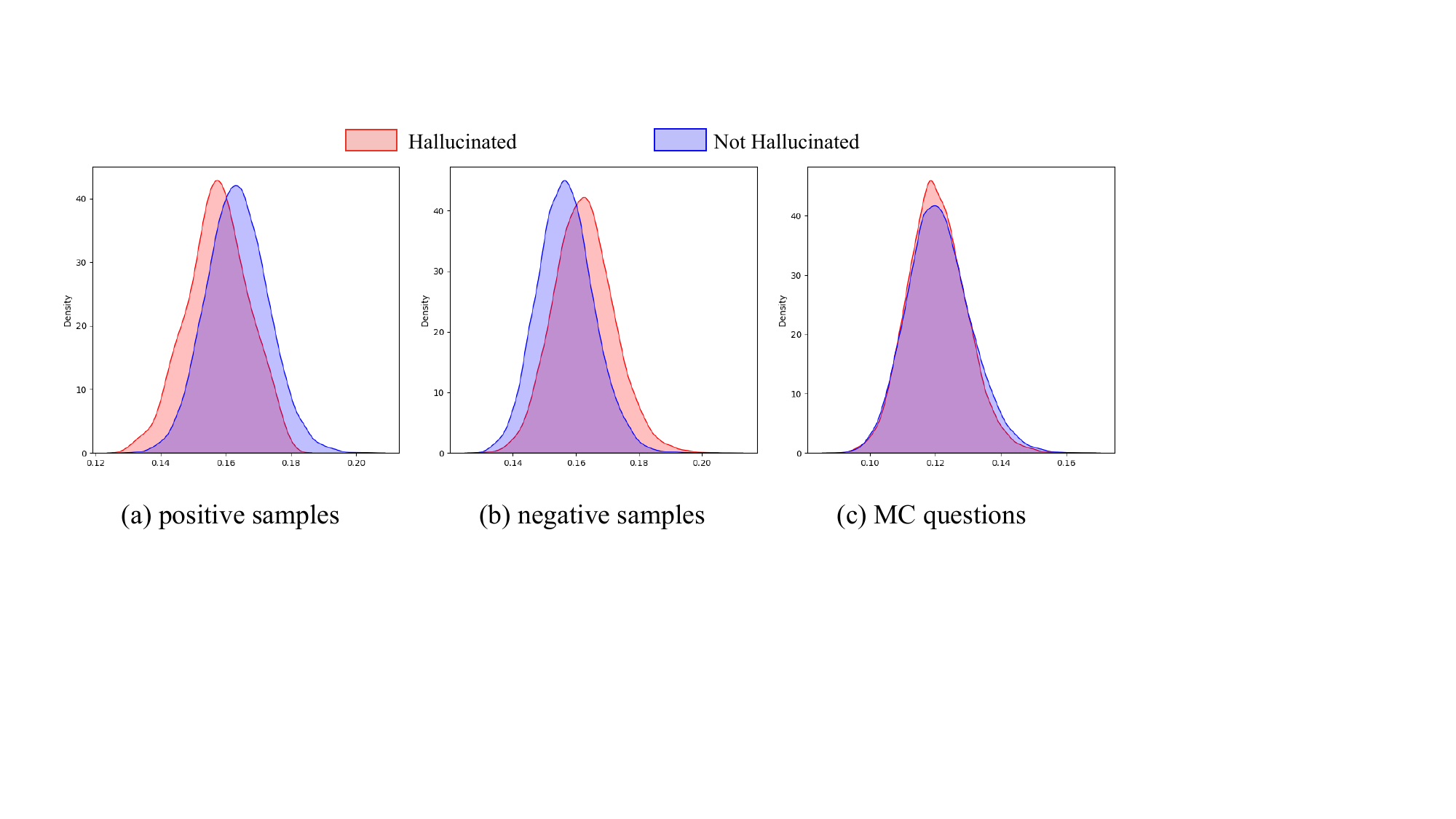}
    \caption{Probability distribution of visual token attention.}
    \label{fig:vmc}
\end{figure}

\textbf{Visual Token Attention}.
\label{vision_token_attention}
The Visual Token Attention is defined as
\begin{gather}
    \mathop{\text{mean}}_j\frac{\sum_{i\in V} \alpha_{ij}}{\sum_{i\in V} \alpha_{ij}+\sum_{k\in T}\alpha_{kj}},
\end{gather}
where $\alpha_{ij}$ represents the attention weight assigned to token $i$ at head $j$ in the last transformer layer, $V$ denotes the set of visual tokens, and $T$ denotes the set of textual tokens. We visualize it in Fig.~\ref{fig:vmc} and we find:
\begin{enumerate}
\item For questions with correct answer \textit{No} hallucinated models tend to give more attention to visual tokens. 
\item For MC questions, there is no obvious difference in vision token attention between hallucinated models and non-hallucinated models.
\end{enumerate}
Therefore, \textbf{unlike what has been found on object hallucination, more attention to visual tokens does not exclude verb hallucination.}


\subsection{Token Uncertainty}
\subsubsection{Token Uncertainty and Hallucination}
\begin{figure}[t]
    \centering
    \includegraphics[width=0.9\linewidth]{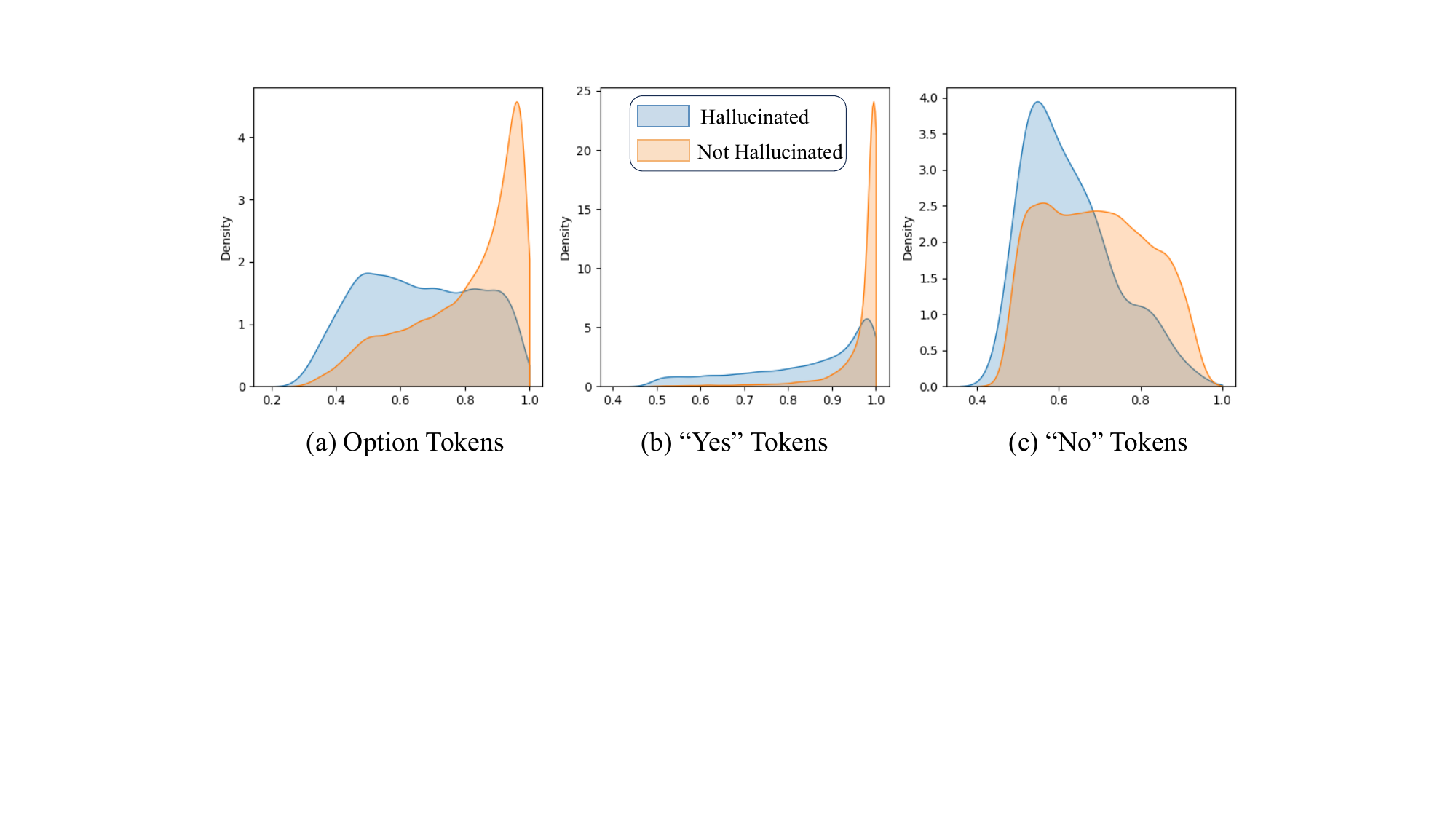}
    \caption{Distribution of token uncertainty.}
    \label{fig:token_uncertainty}
\end{figure}

We dig into the uncertainty~\cite{zhou2023analyzing} of MLLMs via visualizing the distribution of probabilities of predicted tokens of the widely-used open-sourced LLaVA V1.5. 
From Fig.~\ref{fig:token_uncertainty}, we can see that there is a substantial difference between correct answers and hallucinated answers: hallucinated tokens are mostly given with low probability, which means that the model is confused about the answers. 
For Yes-or-No questions, we visualize questions to which the model gives answers \textit{Yes} and \textit{No} separately. The observations are:
\begin{enumerate}
    \item LLaVA V1.5 tends to answer \textit{Yes} with high confidence, but \textit{No} with relatively lower confidence.
    \item Non-hallucinated answers are given with higher confidence than hallucinated ones, regardless of \textit{Yes} and \textit{No}.
    \item For MC questions, answers given by LLaVA V1.5 with higher confidence are very likely to be correct.
\end{enumerate}
In conclusion, MLLM verb hallucination has something in common with MLLM object hallucination. \textbf{Uncertainty is strongly related to MLLM verb hallucination.}
\subsubsection{mAP vs. Acc}
\begin{figure}[t]
\centering
    \includegraphics[width=0.55\linewidth]{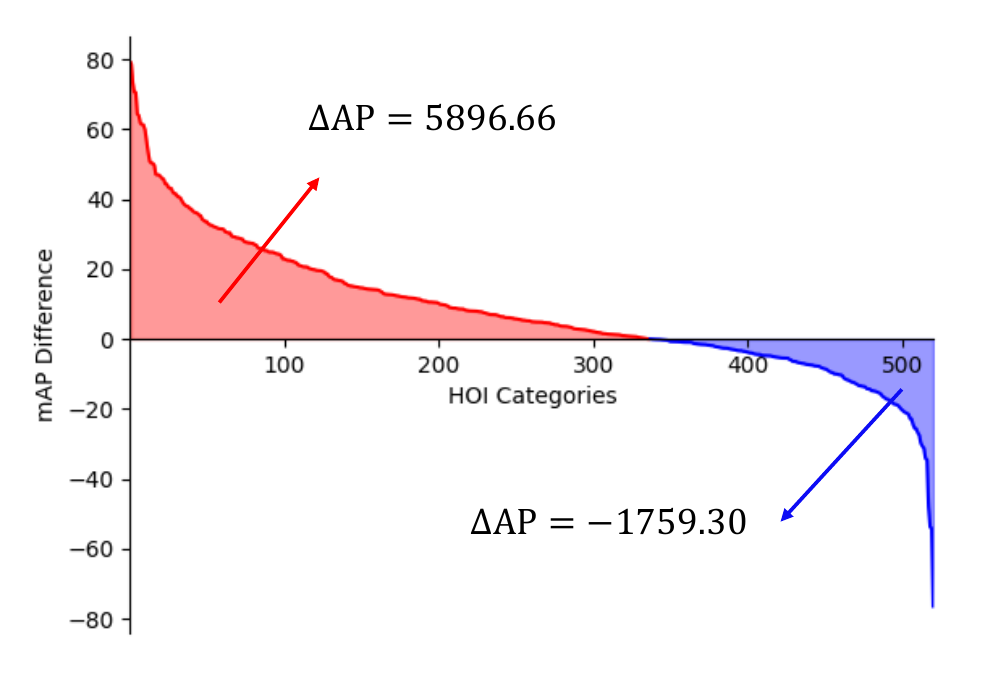}
    \caption{AP Comparison of LLaVA V1.5 and CLIP.}
    \label{fig:map_difference}
\end{figure}

A natural question is, does low accuracy mean poor verb understanding? For example, in Tab.~\ref{object_reference_yon}, LLaVA V1.5 shows an accuracy of 52.16 on YN questions with object references. 
To answer this question, we compare its performance with a strong baseline: HICO-finetuned CLIP~\cite{radford2021learning} with outstanding results~\cite{li2024isolated}. To rule out the influence of objects, we test mAP under the ``Known Object'' setting~\cite{chao2015hico}. We use the probability of \textit{Yes} tokens to calculate mAP. LLaVA V1.5 achieves an mAP of 68.41 while HICO-finetuned CLIP has an mAP of 60.45. The detailed AP gap of each HOI class is in Fig.~\ref{fig:map_difference}. LLaVA V1.5 outperforms CLIP in most HOI classes. This means that LLaVA V1.5, although showing severe verb hallucination, does know well which verbs an image is more likely to contain. Therefore, one of the sources of verb hallucination is the \textbf{mis-calibration of tokens.}

\begin{figure}[t]
    \centering
    \includegraphics[width=0.85\linewidth]{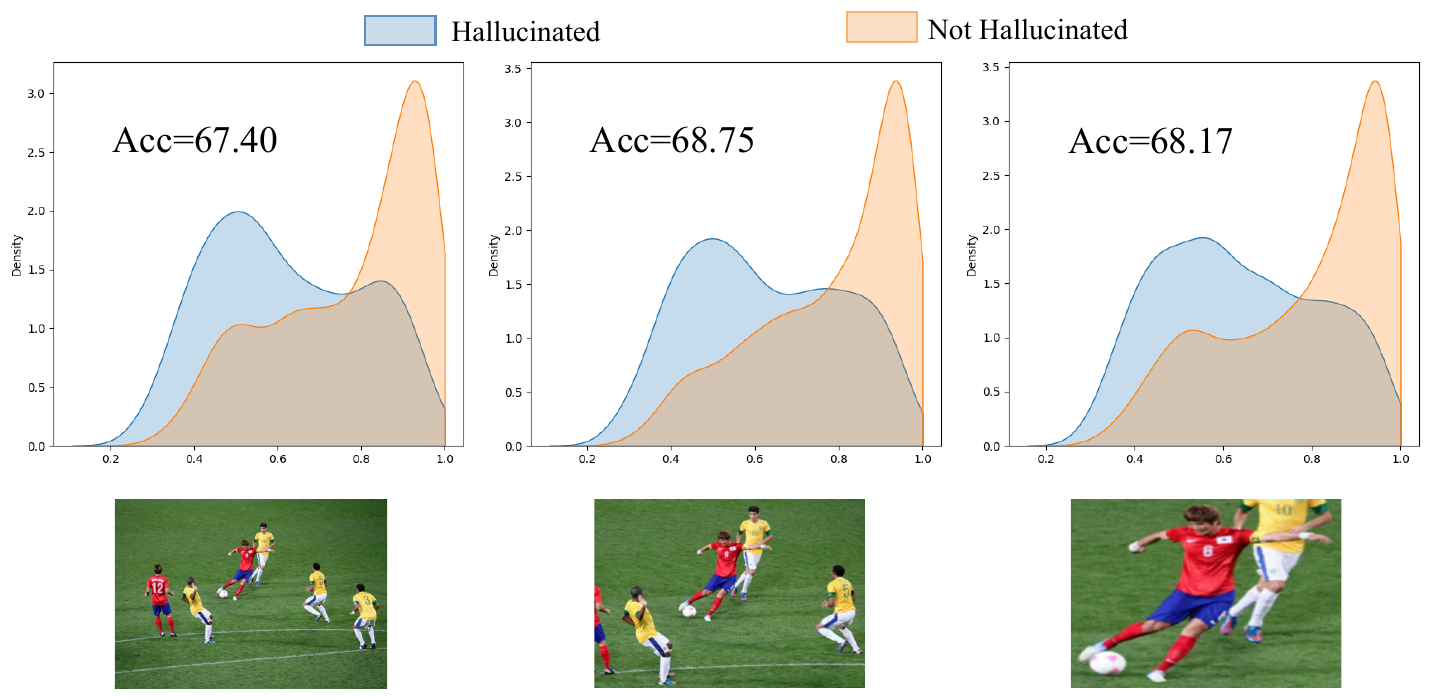}
    \caption{The influence of crop size on token uncertainty.}
    \label{fig:crop_uncertainty}
\end{figure}

\subsubsection{Field of View and Uncertainty}
Previous research~\cite{chen2424halc,zhang2025mllms} has revealed the use of field of view ensembling in hallucination mitigation. \cite{zhang2025mllms} has shown the influence on the change of token uncertainty. We also try to analyze the effect of field of view cropping on MLLM verb understanding. We select from HICO a set of images. For each image, there is only one HOI instance (to rule out the disturbance of multiple instances). The HOI instance is also small enough so that cropped images are substantially different from the original images. The token uncertainty on MC questions visualized in Fig.~\ref{fig:crop_uncertainty} shows that the uncertainty and accuracy do not change much: different from object, verb understanding is not substantially affected by the field of view: MLLMs \textbf{do not always know the verb concepts in the area}.

\section{Hallucination Mitigation Methods}

\subsection{Training-Free Methods}
OPERA~\cite{huang2024opera}, VCD~\cite{leng2024mitigating}, and Nullu~\cite{yang2025nullu} are outstanding training-free hallucination mitigation methods. They do not require fine-tuning and thus are low-cost and have more general applicability.

We present the results of OPERA in Tab.~\ref{hallucination_mitigation}, showing the marginal or even negative effect of OPERA.
Though OPERA tries to punish overreliance on summary tokens and force more attention on visual tokens, we have discussed in Sec.~\ref{vision_token_attention} that more attention to visual tokens does not exclude hallucination.
If the reward is given for attention to visual tokens, hallucination may be worsened. 

VCD regards language prior as a hallucination-inducing factor, uses visual distortion to trigger it, and proposes contrastive decoding to mitigate hallucination. The results of VCD are in Tab.~\ref{hallucination_mitigation}. 
VCD shows inconsistent effects on three models and thus fails on our benchmarks.  
To dig deeper, we compute Qwen-VL's KL divergence between the original token distribution $p_{\theta}(y_t|v,x,y_{<t})$ and contrasted token distribution $p_{vcd}(y_t|v,v',x,y_{<t}))$ of \textbf{20K} samples, and find that the KL divergence of \textbf{18.6K} samples is \textbf{0}, meaning that MLLMs \textbf{rely heavily on language prior to understand verbs, and such prior can not be easily omitted}.

Nullu identifies a low-rank hallucination subspace from truthful and hallucinated responses, projects model weights into its null space to suppress false priors, and reduces hallucination without high cost. Its negative results in Tab.~\ref{hallucination_mitigation} show that \textbf{MLLM layers have not formed reliable distinguishment between truthful and hallucinated verbs}.

\subsection{Influence of Fine-tuning}
REVERIE~\cite{zhang2024reflective}, Haloquest~\cite{wang2024haloquest}, and Octopus~\cite{suo2025octopus} are outstanding fine-tuning methods, and show decent results on object-centric benchmarks.
Octopus train models to adopt different contrastive decoding methods in different cases, while the other two methods fine-tune models with meticulously designed training sets.
However, these training sets do not contain much verb knowledge, and we show their limitation in Tab.~\ref{hallucination_mitigation}. 
To mitigate verb hallucination to some extent
without sacrificing MLLM's ability in other perspectives,
we explore the effect of datasets with rich verb knowledge. 
We try to advance the MLLM with Pangea~\cite{li2024isolated}, which organizes existing heterogeneous action datasets in a unified way. 
It builds a mapping from action labels to abstract verb semantics. 290 frequent verb nodes in VerbNet~\cite{verbnet} are selected and a one-to-290 mapping is built. 
It gives us a whole picture of diverse verbs and carries the structure knowledge of verb relationships.
We select 60K samples from Pangea according to the proportion of the source dataset and build an instruction-tuning dataset. Details are in Suppl. Sec.~{4}. It contains \textbf{280 out of 290} nodes in Pangea P2S mapping and covers a wide range of verb semantics. 
Following common practice~\cite{zhang2024reflective,wang2024haloquest}, we fine-tune LLaVA V1.5 with LoRA. The results are shown in Tab.~\ref{hallucination_mitigation}. Although Pangea only contains \textbf{rough} action labels, it proves more helpful to verb hallucination mitigation than previous training sets. The performance of the original/fine-tuned model on MMMU~\cite{yue2024mmmu}: 32.2/33.0, MathVista~\cite{lu2023mathvista}: 23.6/24.5. The result is not seriously impaired. Our method may also be integrated into MoE in case of interference.
In the future, mining more action data according to the structured verb semantics and fine-tuning MLLMs on them can be a promising way to mitigate verb hallucination.

\begin{table}[t]
\centering
{\scriptsize
\begin{tabular}{ccccccc}
\toprule
                                   & \multicolumn{2}{c}{YN w/ obj} & \multicolumn{2}{c|}{YN w/o obj} &MC w/ obj &MC w/o obj \\
                                   & acc           & F1            & acc            & F1            & acc       & acc       \\
\midrule
LLaVA V1.5                          & 52.16         & 57.69       & 59.16             &\underline{61.79}    & \underline{57.37}&{51.00}\\
OPERA        & 42.46         & 53.69       & 54.45             &59.23    & 57.28     & 51.13\\
VCD      & 52.38         & 58.04       & 57.86             &60.85    & 54.26     & 48.94\\
REVERIE & 40.67         & 52.66         & 41.9           & 53.03        & 37.88 & 41.32    \\
Haloquest & \underline{70.57} & \underline{64.89}&\underline{72.73} &\textbf{63.00} &55.20 & 47.45 \\
Nullu & 51.99 & 57.90 & 59.22 & {61.78} & 55.98 & \underline{53.17} \\
Octopus     & 52.12   & 57.73   &    53.52  &58.71   &  46.59 & 47.55 \\
Ours & \textbf{78.48}& \textbf{68.13} & \textbf{77.37} & {61.61} & \textbf{61.73} &\textbf{60.79} \\
\bottomrule
\end{tabular}}
\caption{Comparison on existing methods.}
\label{hallucination_mitigation}
\end{table}

\section{Conclusion}
In this paper, to our best knowledge, we first reveal MLLM verb hallucination and build a benchmark to probe it from various perspectives. Our experiment reveals that MLLMs suffer from severe verb hallucination in many ways, and existing training-free hallucination mitigation methods fail. Fine-tuning is still the most promising way. However, how to fine-tune existing models efficiently is still a problem to be explored. Experiments show that MLLM verb hallucination is quite different from object hallucination. 
Moreover, whether there are effective training-free verb hallucination mitigation methods is still an open question. 

\section*{Acknowledgements}
This work was supported in part by the National Natural Science Foundation of China under Grant No.~62306175, the Shanghai Committee of Science and Technology, China (Grant No.~24511103200), the National Key Research and Development Project of China (No.~2022ZD0160102), and Shanghai Artificial Intelligence Laboratory, XPLORER PRIZE grants.
{
\small
\bibliography{aaai2026}
}

%% file: supp.tex
\appendix
\twocolumn[
\begin{@twocolumnfalse}
\section*{\centering{Supplementary Material for Verb Mirage: Unveiling and Assessing Verb Concept Hallucinations in Multimodal Large Language Models}}
\end{@twocolumnfalse}
]
\section{Details on Benchmark Construction}
\subsection{Benchmark Based on HICO}
HICO~\cite{chao2015hico} test set contains 9.8K images. We use the HICO test set for benchmark construction. It includes 600 action classes formed by 80 object classes and 117 verb classes. There is a special verb class ``\texttt{no\_interaction}'', and we leave out annotations about it.

Assume $\mathcal{H}$ denotes the set of 600 action classes, $\mathcal{V}$ denotes the set of verb classes, and $\mathcal{O}$ denotes the set of object classes. Each action class $h\in\mathcal{H}$ can be denoted as $(v,o)$ where $v\in V$ and $o\in \mathcal{O}$. 

For MC question construction, given an image $I$ and its HOI annotation $\mathcal{H}_I$, we select $h_0=(v_0,o_0)\in \mathcal{H}_I$ as a positive option, select 3 negative options from the set of plausible but not presented action classes $\{(v,o)\in \mathcal{H}|o=o_0\land (v,o)\notin \mathcal{H}_I\}$, and form a MC question with one correct option. In our benchmark, there are 16.6K unique MC questions.

For YN question construction, we build questions with correct answers \textit{Yes} and \textit{No} similarly to MC question construction. In our benchmark, there are 47K unique YN questions. The ratio of questions with correct answers \textit{Yes} and \textit{No} is approximately 1:2.



\subsection{Benchmark Based on CharadesEgo}

CharadesEgo~\cite{sigurdsson2018actor} contains 157 action classes, each of which is formed by a verb and an object. However, many object classes in CharadesEgo can not match more than 2 verbs. Therefore, to build negative options, we sample from the set $\{(v,o)\in\mathcal{H}|o\in \mathcal{O}_I \land (v,o)\notin \mathcal{H}_I\}$, where $\mathcal{O}_I=\{o|(v,o)\in \mathcal{H}_I\}$. The ratio of questions with correct answers \textit{Yes} and \textit{No} is approximately 1:1. There are 76K unique YN questions and 36K unique MC questions, and we use 28K YN questions and 10K MC questions for experiment in the main paper.

\section{Additional Explanation and Analysis on Model Behaviors}
\subsection{Uncertainty}
\begin{figure}[t]
    \centering
    \includegraphics[width=\linewidth,trim={3cm 4cm 5cm 3cm},clip]{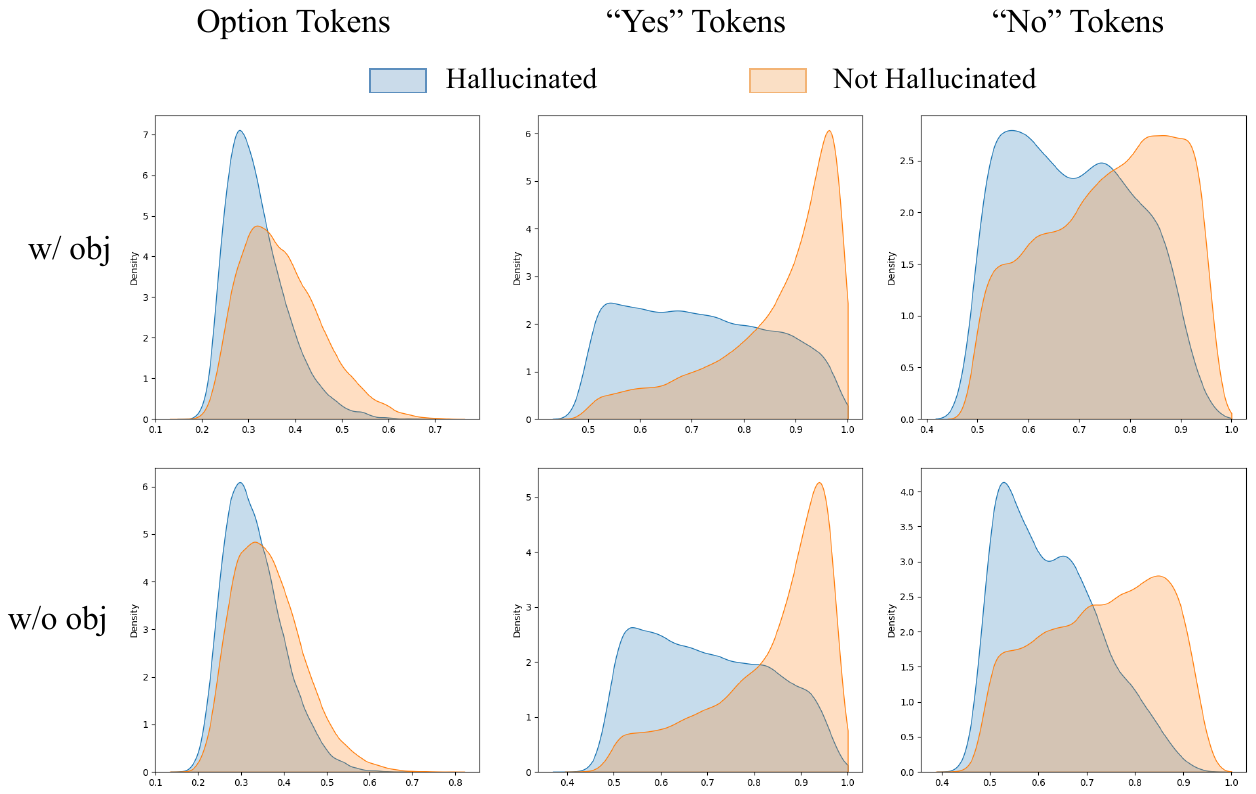}
    \caption{Token uncertainty of InstructBLIP~\cite{dai2023instructblip}. Not hallucinated tokens are often given with high probability.}
    \label{fig:instructblip_uncertainty}
\end{figure}

In the main paper, we visualize token uncertainty:
\begin{equation}
\text{max}\{\text{softmax}(\text{logit}(y_0|v,x))\}
\end{equation}
of LLaVA V1.5~\cite{liu2024improved}. In our setting, $y_0$'s are usually option tokens or \textit{Yes}/\textit{No} tokens, which are used to judge the correctness of MLLMs' answers. Here we provide some additional visualizations in Fig.~\ref{fig:instructblip_uncertainty}. The visualizations give us more solid proof that uncertainty is strongly related to MLLM verb hallucination.
\subsection{Key Image Area Attention}
LLaVA V1.5 adopts CLIP~\cite{radford2021learning} as the visual encoder. 576 tokens generated by CLIP are used as visual tokens and fed to the language model after projection. Each of these 576 tokens represents a small area of the original image. Using the bounding box annotation of HICO, we can get the key image area of a certain verb, represented by the area bounded by the bounding boxes of humans and objects. Our visualization shows that hallucinated MLLM pays less attention to the key image area, but the margin is not large enough. Although MLLM can pay attention to the correct key areas, it can not often understand verb concepts.

\subsection{Verb Token Attention}

Verb Token Attention means the attention given to the verb tokens in question. When visualizing verb token attention, to emphasize LLaVA V1.5's over-attention on noun tokens, we visualize verb token attention normalized by the sum of verb token attention and object token attention:
\begin{equation}
\mathop{\text{mean}}_j\frac{\sum_{i\in T_{v}} \alpha_{ij}}{\sum_{i\in T_{v}} \alpha_{ij}+\sum_{i\in T_{o}}\alpha_{ij}},
\end{equation}
where $\alpha_{ij}$ represents the attention weight assigned to token $i$ at head $j$ in the last transformer layer, $T_{v}$ denotes the set of verb tokens, and $T_{o}$ denotes the set of object tokens.

\subsection{Visual Token Attention}

\begin{figure}
    \centering
    \includegraphics[width=\linewidth,trim={3cm 4cm 5cm 3cm},clip]{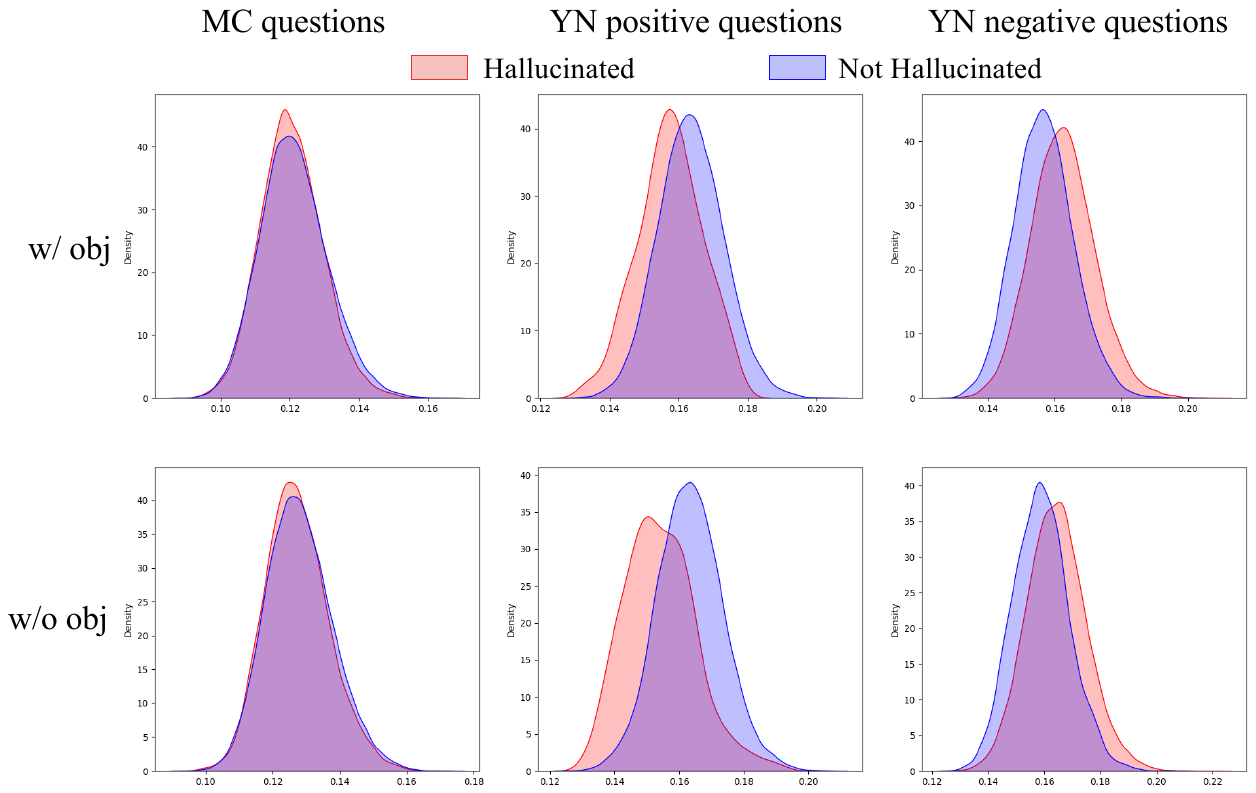}
    \caption{Visual token attention distribution for LLaVA V1.5.}
    \label{fig:llava_vmc}
\end{figure}

\begin{figure}
    \centering
    \includegraphics[width=\linewidth,trim={3cm 4cm 5cm 3cm},clip]{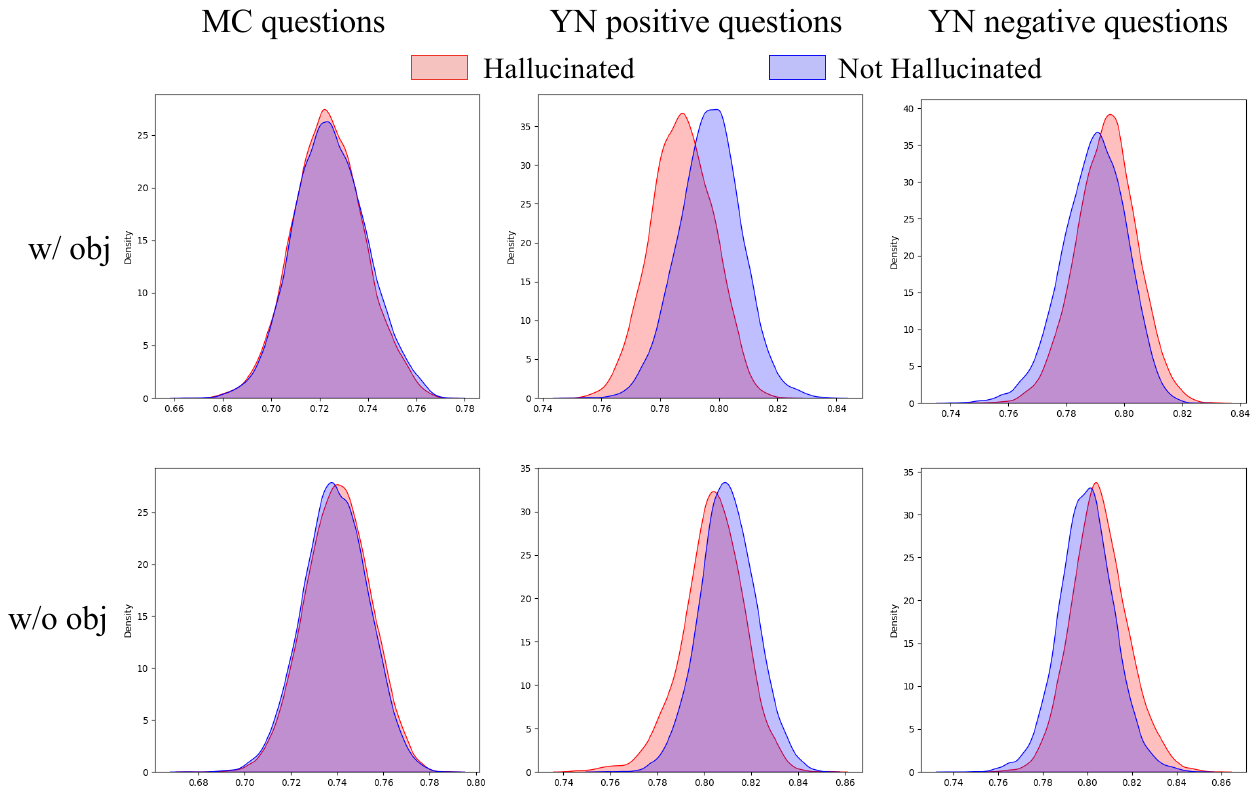}
    \caption{Visual token attention distribution for InstructBLIP.}
    \label{fig:instructblip_vmc}
\end{figure}

Visual Token Attention(VTA) is defined similarly to Visual Modality Contribution(VMC)~\cite{chen2024multi}. The Visual Token Attention is defined as
\begin{gather}
    \mathop{\text{mean}}_j\frac{\sum_{i\in V} \alpha_{ij}}{\sum_{i\in V} \alpha_{ij}+\sum_{k\in T}\alpha_{kj}},
\end{gather}
where $\alpha_{ij}$ represents the attention weight assigned to token $i$ at head $j$ in the last transformer layer, $V$ denotes the set of visual tokens, and $T$ denotes the set of textual tokens.
In the implementation of OPERA~\cite{huang2024opera}, when generating the first 10 tokens, a reward is given for high visual token attention. We visualize VTA on other models and other question types in Fig.~\ref{fig:llava_vmc} and ~\ref{fig:instructblip_vmc}, and have the following observations:
\begin{enumerate}
    \item For questions with correct answers \textit{Yes}, hallucinated models tend to give less attention to visual tokens.
    \item For questions with correct answers \textit{No}, hallucinated models tend to give more attention to visual tokens.
    \item For MC questions, there is no substantial difference in visual token attention.
\end{enumerate}

The observation above can explain the reason for the marginal or inconsistent effect of OPERA on MLLMs.

\subsection{Object Reliance}
\begin{figure}[h]
    \centering
    \includegraphics[width=\linewidth]{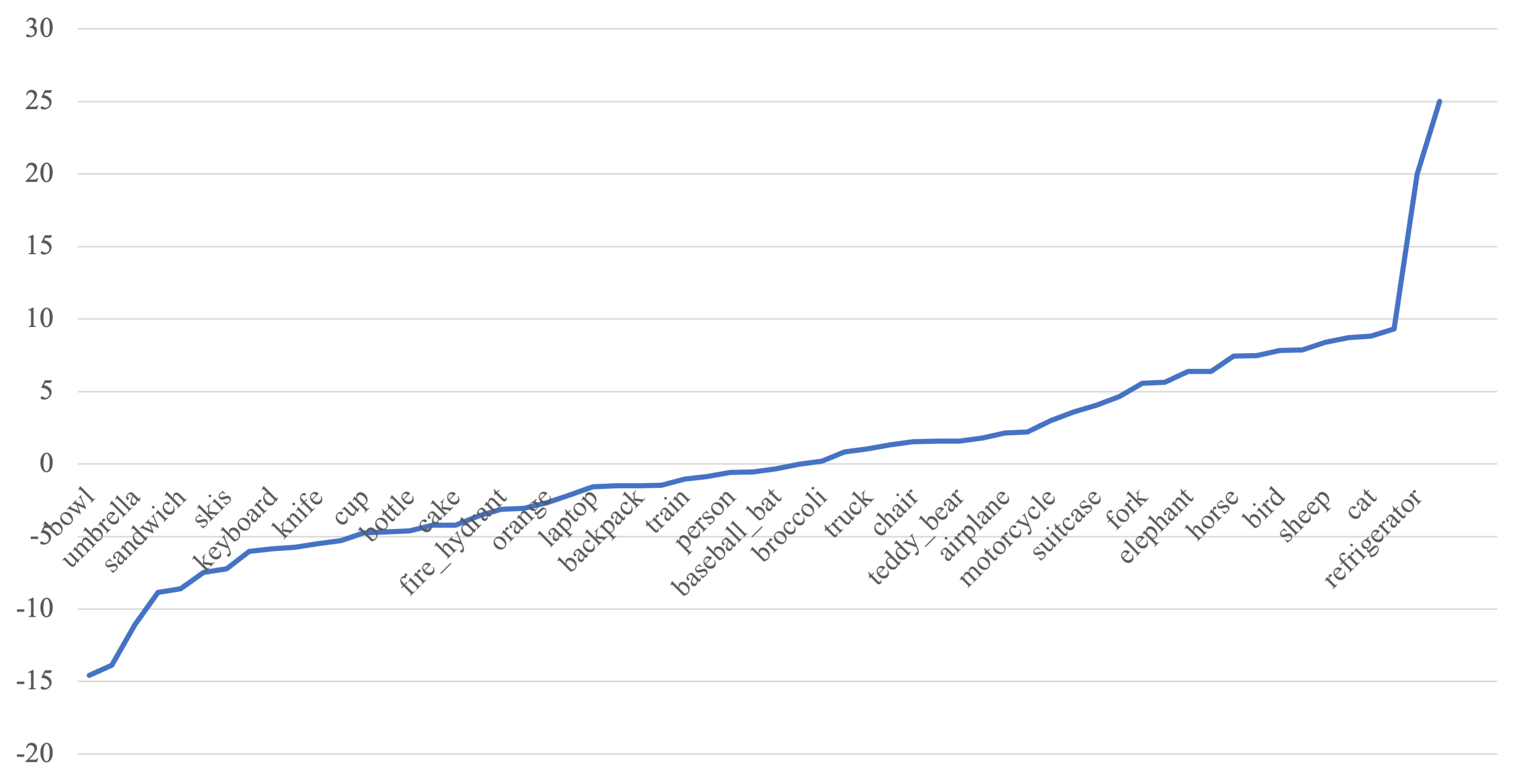}
    \caption{mPLUG-Owl2's difference of YN accuracy with and without object references.}
    \label{fig:acc_diff}
\end{figure}

We split YN question sets according to object classes, and show the difference between mPLUG-Owl2's~\cite{ye2024mplug} accuracy with and without object references in Fig.~\ref{fig:acc_diff}. We can see that for questions about most object classes, mPLUG-Owl2 behaves differently with and without object references and thus is affected by object references although the accuracy does not change much (62.94/62.61).

\begin{table}[t]
\centering
\resizebox{0.85\linewidth}{!}{
\begin{tabular}{ccccc}
\toprule
     & Qwen-VL-Chat & mPLUG-Owl2 & LLaVA V1.5  & InstructBLIP   \\
\midrule
overlap  & 95.35    & 79.17  & 82.14      & 100           \\
\bottomrule
\end{tabular}}
\caption{Influential object set overlap of MLLMs.}
\label{influential_object_set_overlap}
\end{table}
\begin{table}[t]
\centering
\resizebox{0.85\linewidth}{!}{
\begin{tabular}{ccccc}
\toprule
IoU           & Qwen-VL-Chat  & LLaVA V1.5  & mPLUG-Owl2  & InstructBLIP \\
\midrule
Qwen-VL-Chat  &   -           & 59.30      &  42.86  & 52.78        \\
LLaVA V1.5     & 59.30         & -          &  46.91  & 53.57        \\
mPLUG-Owl2        &  42.86     & 46.91   & -       &    41.54     \\
InstructBLIP  & 52.78         & 53.57      & 41.54   & -           \\
\bottomrule
\end{tabular}}
\caption{YN influenced verb set IOU.}
\label{Verb_IOU}
\end{table}

\subsection{MLLM Error Consistency}
\textbf{Consistency on MC and YN questions.} 
From main paper Tab.~{\color{red}1}, we find that there is inconsistency in MLLMs' performance on multiple-choice and yes-or-no tasks. However, it is an interesting question whether there is some consistency. 
Thus, we divide the questions into different classes according to object class references and get the object classes that influence MLLMs the most. 
Assume $S_1$ denotes the set of object classes without which MLLMs can not recall correct verbs given YN questions, and $S_2$ denotes the similar set of object classes for MC questions. 
We define overlap index=$\frac{|S_1\cap S_2|}{\text{min}\{|S_1|,|S_2|\}}$, and compute MLLMs' overlap index. From Tab.~\ref{influential_object_set_overlap}, we find substantial overlap between $S_1$ and $S_2$. A substantial number of object classes can affect MLLM verb understanding of both question types.

\textbf{Consistency among models.} 
We analyze models and get verb classes that are influenced by object references most according to per verb class \textit{Yes} ratio. 
From Tab.~\ref{Verb_IOU}, we can see that there is substantial overlap among different models. 



\section{Details about Experiments on Training-free Hallucination Mitigation}
\subsection{Experiment Setting}

Previous research~\cite{leng2024mitigating} has demonstrated that sampling strategy can affect MLLM performance. Therefore, to evaluate the effect of hallucination mitigation methods fairly, we choose to control the use of decoding strategy. As OPERA~\cite{huang2024opera} heavily relies on beam search, we select beam search as the default decoding strategy for evaluation on OPERA, VCD, and other methods in the main paper. All experiments are conducted using a server with 2 A100 40G GPUs.

We test VCD with sampling as the decoding strategy as the author claimed in the paper~\cite{leng2024mitigating}, and get results in Tab.~\ref{VCD}. From the result, we can see that VCD also shows inconsistent results. Sometimes sampling+VCD may achieve substantial improvement over sampling only, but the result is not better than beam search only.

\begin{table*}[!p]
\resizebox{\linewidth}{!}{
\begin{tabular}{ccccc|ccc}
\toprule
                           &             & \multicolumn{3}{c|}{w/ obj}               & \multicolumn{3}{c}{w/o obj}              \\
\midrule
                           &             & Qwen-VL-Chat & LLaVA V1.5 & InstructBLIP & Qwen-VL-Chat & LLaVA V1.5 & InstructBLIP \\
\multirow{3}{*}{YN acc}    & sampling      & \underline{77.55}        & \underline{51.46}      & 65.96        & \underline{78.94}        & 52.21      & 65.34        \\
                           & sampling+VCD  & 76.68         &52.14      & \underline{68.59}        & 78.79        & \underline{53.12}      & \underline{68.67}        \\
                           & beam search & \textbf{78.06}& \textbf{52.16}       & \textbf{72.53}        & \textbf{79.24}        & \textbf{59.16}      & \textbf{73.82}        \\
\midrule
\multirow{3}{*}{YN prec}   & sampling      & \underline{61.63}& 40.56      & 49.5         & 64.49        & 41.09      & 49.08        \\
                           & sampling+VCD  & 60.72        & \underline{40.98}    & \underline{52.02}        & \textbf{65.34}        & \underline{41.64}      & \underline{52.28}        \\
                           & beam search & \textbf{62.37}        & \textbf{40.99}       & \textbf{55.79}        & \underline{65.09}        & \textbf{45.10}      & \textbf{57.25}        \\
\midrule
\multirow{3}{*}{YN recall} & sampling      & \textbf{87.4}         & \underline{96.37}      & 78.65        & \textbf{83.29}        & 96.73      & 77.98        \\
                           & sampling+VCD  & 86.06        & 97.35      & \underline{80.52}        & 78.79        & \underline{97.74}      & \underline{79.59}        \\
                           & beam search & \underline{87.02}        & \textbf{97.35}      & \textbf{86.77}        & \underline{82.68}        & \textbf{98.06}      & \textbf{87.79}        \\
\midrule
\multirow{3}{*}{MC acc}    & sampling      & \underline{55.74}        & 37.15      & 0.9          & \underline{54.26}        & 39.29      & 0.73         \\
                           & sampling+VCD  & 55.74        & \underline{46.41}      & \underline{1.55}         & 52.09        & \underline{47.47}      & \underline{1.01} \\
                           & beam search & \textbf{55.95}        &\textbf{57.37}      & \textbf{13.48}        & \textbf{54.57}        & \textbf{51.00}         & \textbf{6.25}     \\
\bottomrule
\end{tabular}}
\caption{Comparison on sampling, sampling+VCD, and beam search. \textbf{Bold}: best among three. \underline{Underline}: second best among three.}
\label{VCD}
\vspace{30px}
\end{table*}

\section{Details about Mitigation via Pangea Fine-tuning}

Pangea~\cite{li2024isolated} gathers many heterogeneous datasets in a unified way. We omit Kinetics-700~\cite{carreira2019short} which makes too large a proportion in Pangea, and HAKE~\cite{li2022hake} which is in the same domain as HICO. Among other source datasets, we sample 60K images according to the proportion they make up in Pangea. In this way, we sample massive out-of-domain images, each of which only contains rough action labels. 

We transform these data into MC and YN questions. When building MC questions, for an image and its action label, we sample wrong action labels from its source dataset. The prompt looks like ``Which of the following actions is present at the image? A. [action A]$\backslash$n B. [action B]$\backslash$n C. [action C]$\backslash$n D. [action D]'' When building YN questions, for an image and its action label, to build a negative question, we also sample wrong action labels from its source dataset. The prompt looks like ``Does this image show a person's activity: [action]?''

We gather MC questions and YN questions together. There are 60K MC questions and 60K YN questions. The ratio of YN questions with correct answers \textit{Yes} and \textit{No} is approximately 1:19. The learning rate is 2e-5. Training is conducted using a server with 2 A100 40G GPUs with a total batch size of 64.

\section{Failure Case Visualization}
\begin{figure*}[!p]
    \centering
    \vspace{-50px}
    \includegraphics[width=\textwidth,trim={{1cm} {3cm} {1.5cm} {4cm}},clip]{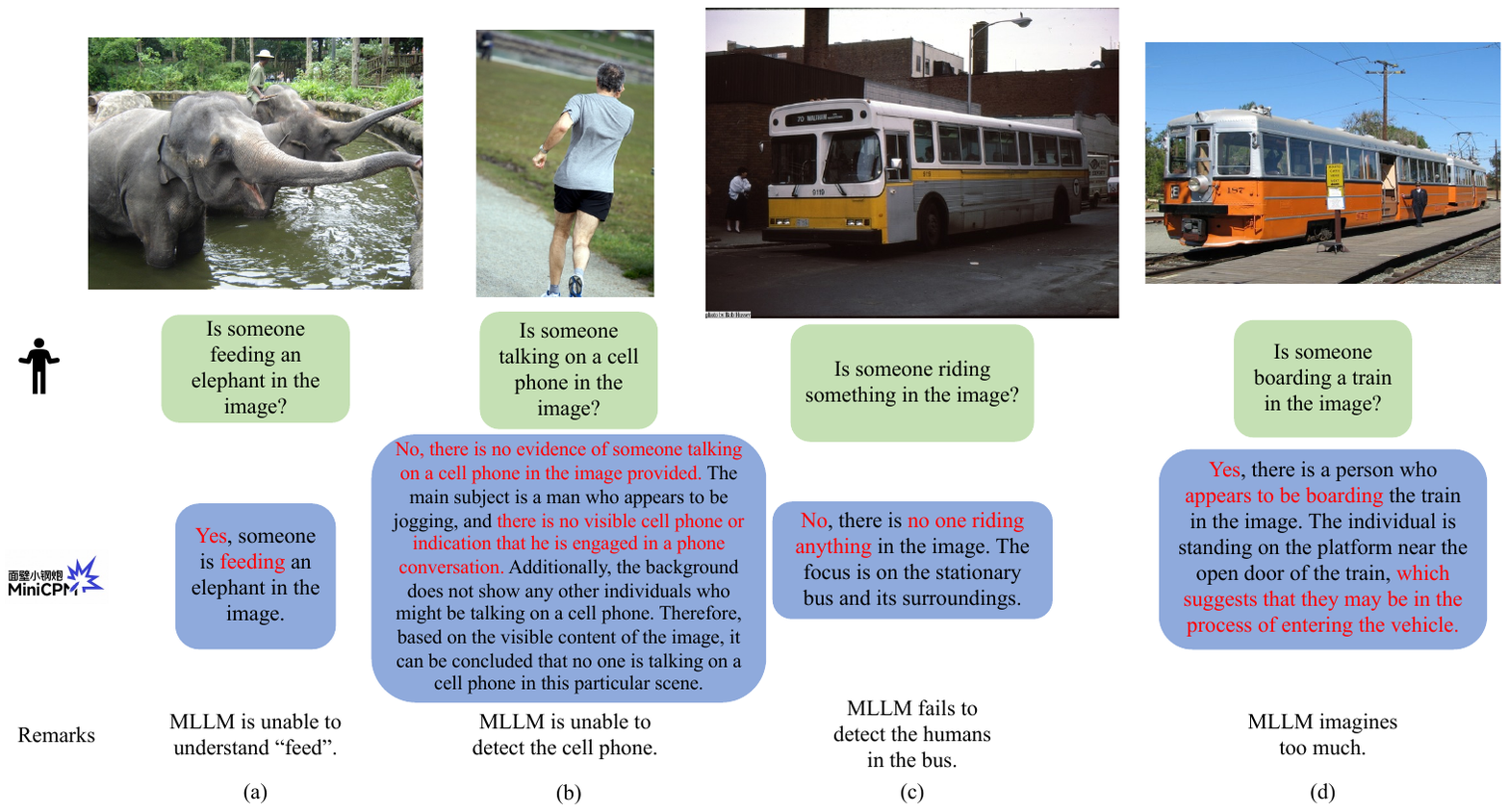}
    \vspace{-20px}
    \caption{Some failure cases of MiniCPM-Llama3-V2.5~\cite{yao2024minicpm}.}
    \label{fig:minicpm_failure}
    \vspace{-20px}
\end{figure*}

\begin{figure*}[!p]
    \centering
    \vspace{15px}
    \includegraphics[width=\textwidth,trim={{1cm} {3cm} {1.5cm} {4cm}},clip]{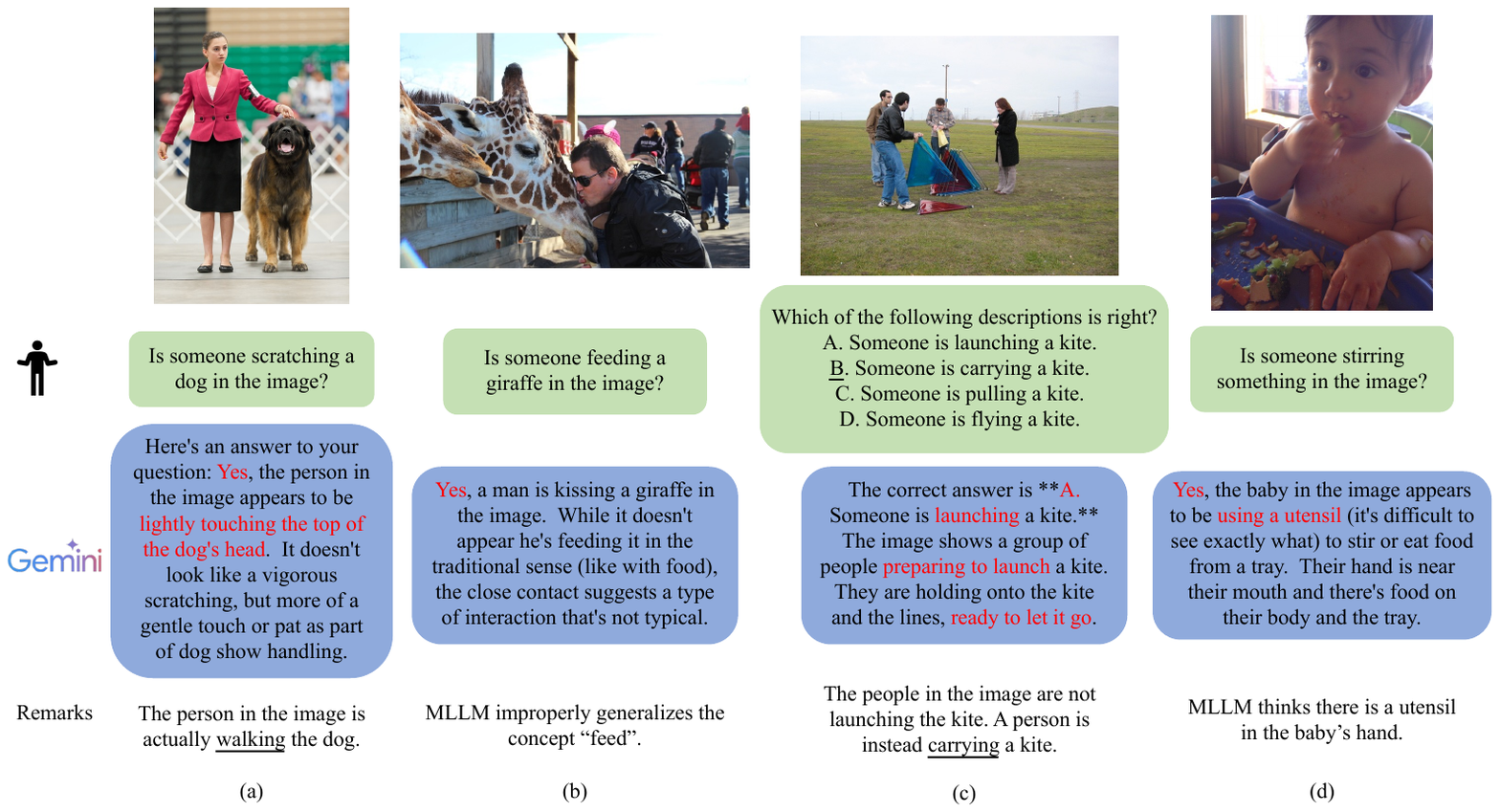}
    \vspace{5px}
    \caption{Some failure cases of Gemini-1.5-Flash~\cite{team2023gemini}.}
    \label{fig:gemini_failure}
\end{figure*}

\begin{figure*}[!p]
    \centering
    \vspace{-50px}
    \includegraphics[width=\textwidth,trim={{1cm} {3cm} {1.5cm} {4cm}},clip]{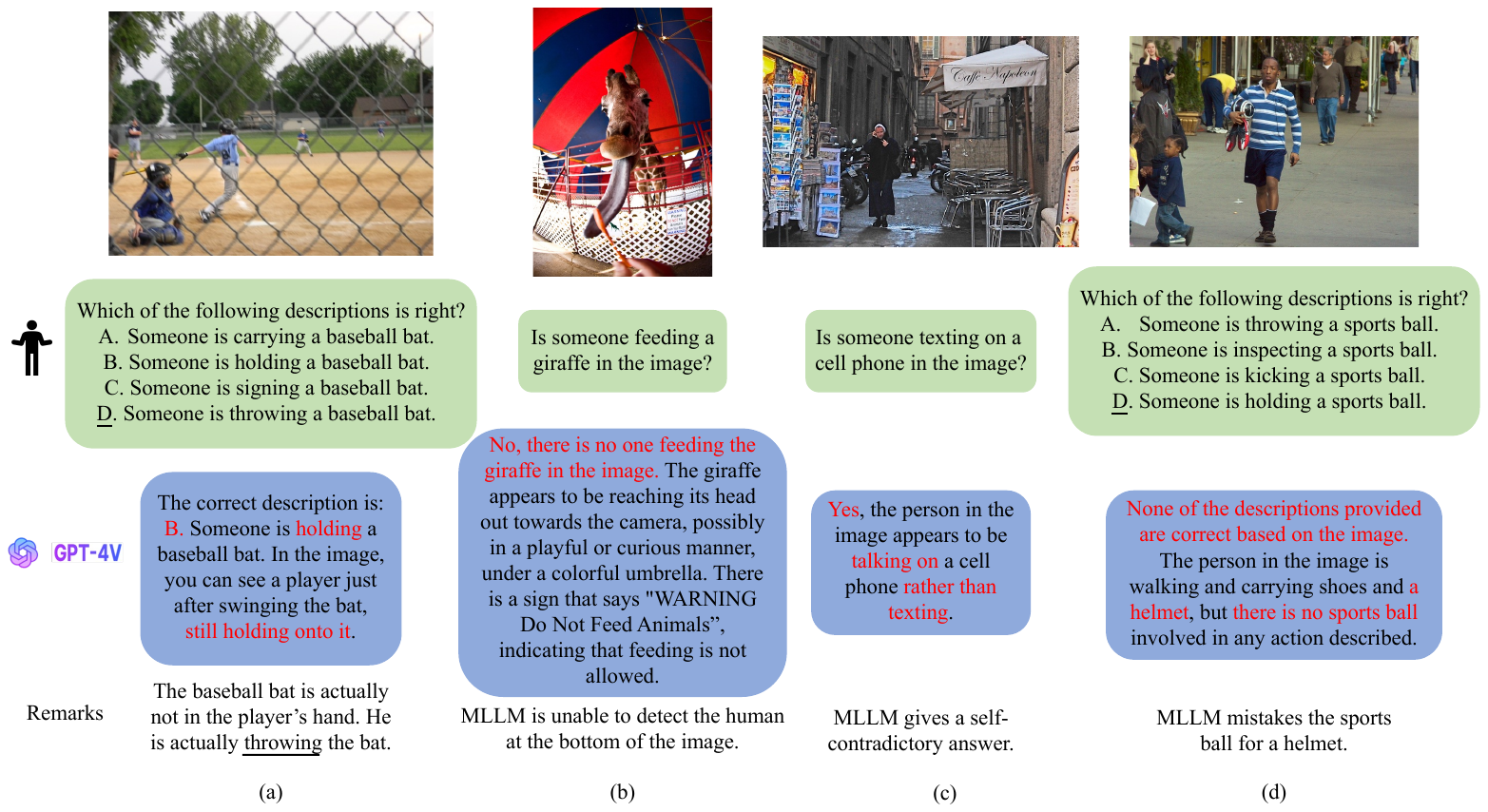}
    \vspace{-20px}
    \caption{Some failure cases of GPT-4-Turbo~\cite{achiam2023gpt}.}
    \label{fig:GPT4_failure}
    \vspace{-20px}
\end{figure*}

In this section, we showcase some samples of which MLLMs hallucinate. MLLMs' responses in {\color{red}red} denote hallucinations. We can see MLLMs hallucinate because of inability to detect humans (Fig.~\ref{fig:minicpm_failure}(c) and Fig.~\ref{fig:GPT4_failure}(b)), inability to understand objects (Fig.~\ref{fig:minicpm_failure}(b), Fig.~\ref{fig:gemini_failure}(d), and Fig.~\ref{fig:GPT4_failure}(d)), and inability to distinguish verbs (Fig.~\ref{fig:minicpm_failure}(a)(d), Fig.~\ref{fig:gemini_failure}(a)(b)(c), and Fig.~\ref{fig:GPT4_failure}(a)(c)). A detailed analysis of MLLMs' chains of thought can reveal that MLLMs can sometimes generate self-contradicting responses (Fig.~\ref{fig:GPT4_failure}(c)), imagine too much (Fig~\ref{fig:minicpm_failure}(d)), or improperly generalize concepts (Fig.~\ref{fig:gemini_failure}(b)), which all lead to MLLM verb hallucination.

\section{More Discussion}
In this paper, we unveil, assess, and study the verb hallucination problem of MLLMs. Many fields such as human-robot interaction, sports analysis, \textit{etc.} require deep understanding of action verbs, and have images captured under similar fields of view as input. Therefore, our study can inspire broader scopes. 

 As the first step, we test many MLLMs as well as outstanding methods targeted object hallucination. In addition, we also tried some ad-hoc methods, such as adding ``Pay attention to the verb in question.'' to the prompt: the F1 score of LLaVA 1.5 on YN questions is 57.5 while that on YN questions without object references is 62.4, showing hardly any gain. To our best knowledge, verb hallucination is still a much harder problem than object hallucination. As for Qwen2.5-VL-7B, the F1 score of YN and YN without object references are 73.79 and 71.99 respectively. Compared with its F1 score of 85.9 on POPE, its verb hallucination is still severer. We have also tried to analyze the influence from network architectures, but were hindered by the fact that mainstream MLLMs are trained with private data, which renders impossible an apple-to-apple comparison. The field of image generation has seen the emergence of Diffusion models based on noise prediction~\cite{rombach2022high} and rectified flow~\cite{liu2023flow}, and Visual Autoregressive Models~\cite{tian2024visual}, and we hope similar exploration and wide adoption of paradigm shift in mainstream MLLMs.

